\pdfoutput=1

\documentclass[11pt]{article}

\usepackage{EACL2023}

\usepackage{times}
\usepackage{latexsym}
\usepackage[T1]{fontenc}
\usepackage[utf8]{inputenc}
\usepackage{microtype}
\usepackage{inconsolata}

\usepackage{booktabs}
\usepackage{graphicx}

\def\aclpaperid{531} 

\title{\textit{You Know What You Need}: Guiding Large Language Models for Social Conversation Synthesis Using In-Context Examples}

\title{\textit{PLACES}: Prompting Language Models for Social Conversation Synthesis}

\author{Maximillian Chen$^1$\thanks{~~Work done during internship at Amazon Alexa AI}~, 
  Alexandros Papangelis$^{2}$, 
  Chenyang Tao$^{2}$,
  Seokhwan Kim$^2$, \\
  \textbf{Andy Rosenbaum}$^2$, 
  \textbf{Yang Liu}$^2$, 
  \textbf{Zhou Yu}$^1$,
  \textbf{Dilek Hakkani-Tur}$^2$\\
  $^1$Columbia University, $^2$Amazon Alexa AI \\
  \texttt{maxchen@cs.columbia.edu, zy2461@columbia.edu} \\
  \texttt{\{papangea,chenyt,seokhwk,andros,yangliud,hakkanit\}@amazon.com}
}

\date{}

\begin{document}
\maketitle

\begin{abstract}
Collecting high quality conversational data can be very expensive for most applications and infeasible for others due to privacy, ethical, or similar concerns. 
A promising direction to tackle this problem is to generate synthetic dialogues by prompting large language models. 
In this work, we use a small set of expert-written conversations as in-context examples to synthesize a social conversation dataset using prompting
\footnote{
\url{https://github.com/alexa/PLACES}
}.
We perform several thorough evaluations of our synthetic conversations compared to human-collected conversations. This includes various dimensions of conversation quality with human evaluation directly on the synthesized conversations, and interactive human evaluation of chatbots fine-tuned on the synthetically generated dataset. 
We additionally demonstrate that this prompting approach is generalizable to multi-party conversations, providing potential to create new synthetic data for multi-party tasks. Our synthetic multi-party conversations were rated more favorably across all measured dimensions compared to conversation excerpts sampled from a human-collected multi-party dataset. 
\end{abstract}

\section{Introduction}
Training dialogue models typically requires an abundance of data, as with any machine learning task. However, collecting high quality data is difficult and expensive, especially for dialogue tasks where there often is no ``right answer'' when developing the trajectory of a conversation. Typically dialogue data are sourced from crowdworkers and the quality of annotations, evaluations, and conversations can vary considerably \cite{zhao2014evaluation}, often necessitating guardrails such as credential-based worker selection or defensive task design for quality control~\cite{allanbaksh2013quality}. 
\begin{figure}[h]
    \centering
    \includegraphics[width=\linewidth]{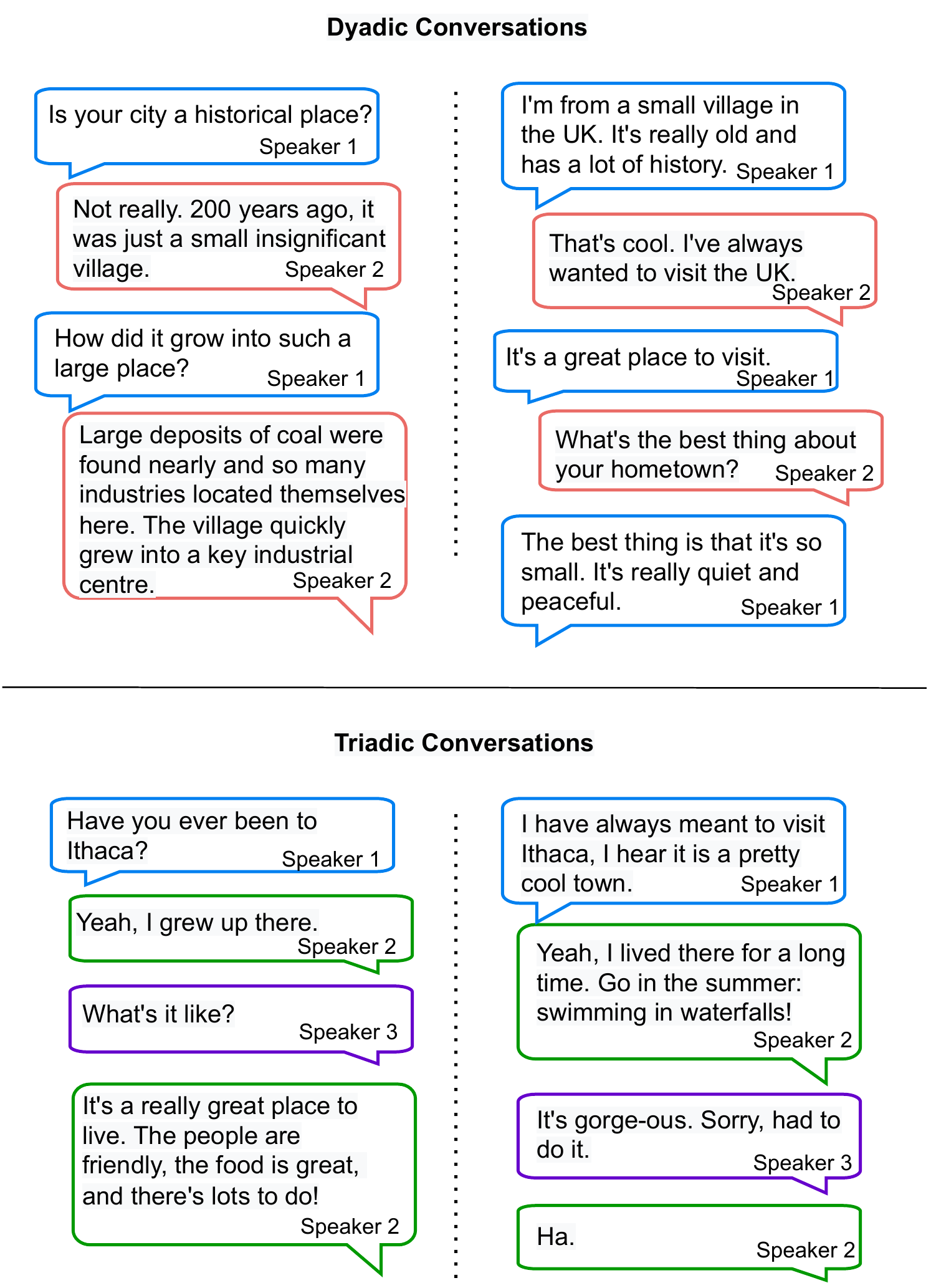}
    \caption{Pair of dyadic conversation excerpts about hometowns (upper) and pair of triadic conversation excerpts about Ithaca, NY (lower). In both pairings, one conversation is synthetically generated and the other is collected from humans. The answer is in Section~\ref{conversation_evaluation}.}
    \label{fig:Examples}
    \vspace{-4mm}
\end{figure}

To accommodate data scarcity in training dialogue tasks, low resource methods have become a topic of growing interest and importance  \citep{zhao2019low,mi2019meta,qian2019domain,li2019insufficient}. One idea that has gained particular attention is transfer learning --- specifically, finding ways to leverage knowledge learned by pre-trained large language models (PLMs) for new tasks. PLMs have demonstrated impressive emerging conversational capabilities, enabling big performance improvements in various dialogue tasks~\cite{brown2020language,shuster2022blenderbot,peng2022godel,kulhanek2021augpt}. Particularly, PLMs have been prompted to augment existing conversational data  \citep{chen2022weakly,mehri2022LAD,sahu2022data}. 

\begin{figure*}
    \centering
    \includegraphics[width=0.9\linewidth]{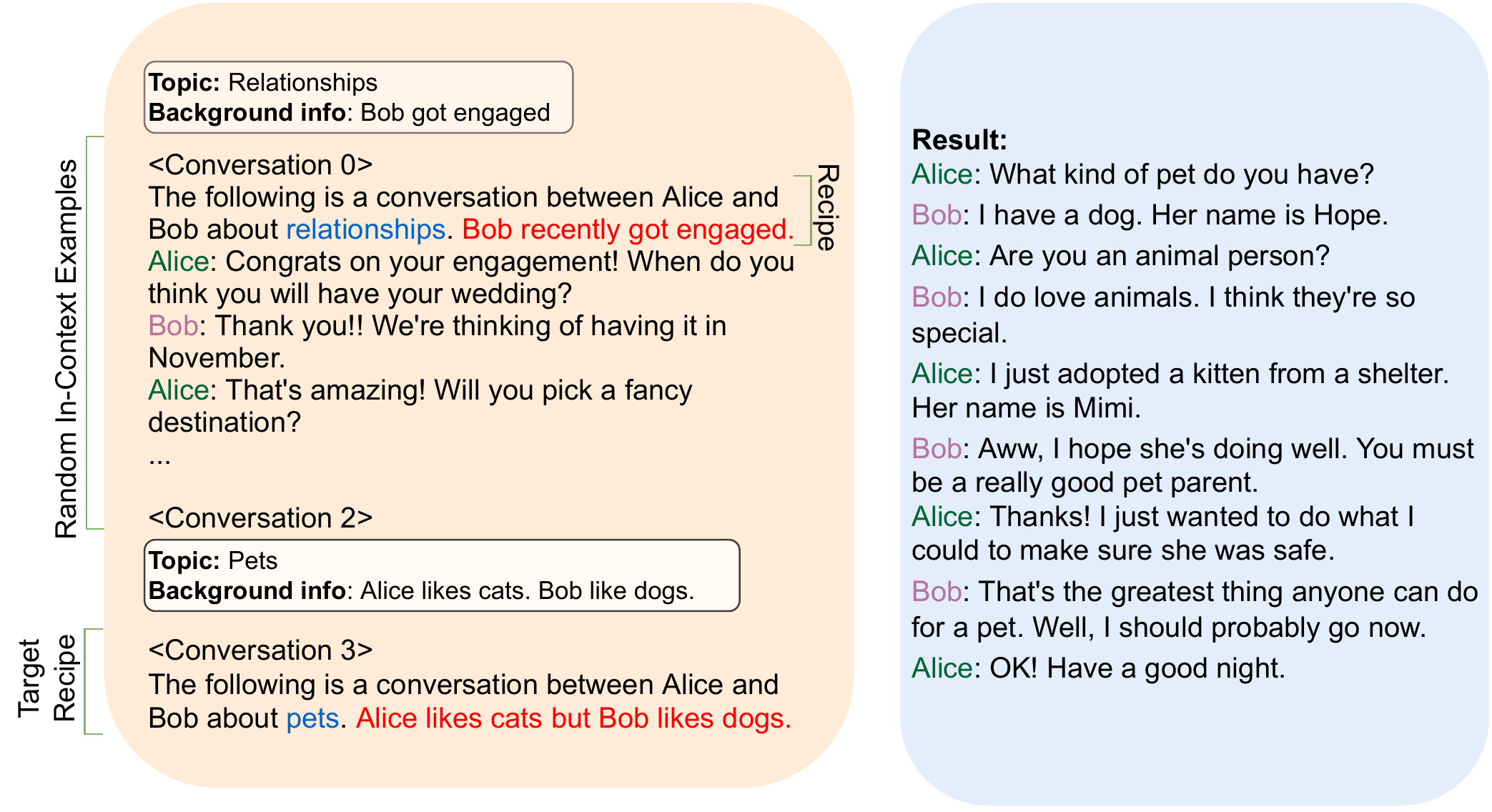}
    \caption{Example of the components of a prompt (left) used by OPT 30B to generate a synthetic conversation about pets (right). Conversations in the prompt are prefixed by recipes. Blue text: topic labels. Red text: seed background information metadata.}
    \label{fig:prompt_example}
    \vspace{-4mm}
\end{figure*}

Given some in-distribution seed examples, augmentation techniques attempt to generate data that are faithful to some task distribution~\cite{kim2021linda}. 
Albeit powerful, one caveat common to all augmentation techniques is that the quality of synthetic data heavily relies on seed examples.
But, what if crowdworkers do not possess the necessary background or skill set to complete a task en masse? How can we still get adequate high-quality synthetic data to learn a task? 

In this work, we explore a novel application of {\bf P}rompting {\bf LA}nguage models for social {\bf C}onv{\bf E}rsation {\bf S}ynthesis (PLACES). Synthesizing conversational datasets allows for the construction of training instances in nonexistent tasks. 
We specifically conduct open-domain, topic-conditioned conversation generation using few-shot in-context learning with expert-written synthetic conversations. We conjecture that expert end-users know exactly the types of conversations that they need. Rather than using existing datasets, they can simply write a small set of high quality conversation examples according to the structure of their desired conversational outputs. 
We reason that given structure through high-quality in-context demonstrations, large PLMs are able to utilize their expansive pre-training data (e.g. \citet{gao2020pile}) to synthesize realistic social conversations, implicitly creating personalities and backgrounds for hypothetical speakers. The process of conversation writing would otherwise require human creativity and effort.

Our paper makes four core contributions. \\(1) 
PLACES involves synthesizing
an entire conversational dataset from a few targeted expert-written examples. These conversations match the quality of two widely adopted social dialogue datasets, DailyDialog~\cite{li2017dailydialog} and Topical Chat~\cite{gopalakrishnan2019topical}, in terms of human evaluation and automatic metrics. 
(2) We demonstrate that our synthetic conversations can be used as a fine-tuning dataset which matches the performance of its human-curated counterparts as measured by an interactive human evaluation and automatic metrics.
(3) We apply PLACES to synthesize data for an under-studied subfield of dialogue research: multi-party conversations. 
We evaluate a set of synthetic triadic conversations in comparison to two human-collected multi-party conversational datasets~\cite{shaikh2010mpc,poria2019meld}. To our knowledge, our work is the first to synthesize multi-party conversations, adding to the still-growing body of work on multi-party social dialogue. (4) Lastly, we conduct an error analysis on both dyadic and triadic synthetic conversations. We discuss the implications of our findings, as well as potential solutions to address the generation ``errors.''

\section{Related Work}
 Recently, the zero- and few-shot learning capabilities of large pre-trained language models have overtaken state-of-the-art performance on many classical natural language processing tasks, including dialogue~\cite{brown2020language}. Many PLMs such as T5~\cite{raffel2020exploring}, GPT-J~\cite{gpt-j}, GPT-3~\cite{brown2020language}, and OPT~\cite{zhang2022opt} have become the backbone of several dialogue-specific models (e.g., \citet{peng2022godel,madotto2021few,shuster2022blenderbot}).

In particular,  in-context learning, where few-shot examples are provided in the input prompt of a PLM, has been found to provide valuable information in guiding generation output \cite{min2022rethinking, brown2020language,min2021metaicl,lu2021fantastically}. As a result, many recent efforts in prompting PLMs have sought to augment various natural language processing datasets \cite{chen2022weakly,wang2022promda,sahu2022data,mehri2022LAD,rosenbaum2022clasp}. Prompting has become a viable ``solution'' for augmentation in dialogue tasks, which have traditionally been considered challenging due to the difficulty of augmenting dialogue context~\cite{chen2022weakly}. 

However, prompt-based augmentation strategies are uncontrolled forms of generation, which may result in generation mistakes for labeled datasets~\cite{sahu2022data,chen2022weakly,meng2022generating}.
In contrast, other recent studies have instead proposed language augmentation strategies that use complex, highly-controlled frameworks that often involve fine-tuning generators \cite{Papangelis2021GenerativeCN,Zhang2020DialogueDO,kulhanek2021augpt,Zhang2020GroundedCG}. Such complex augmentation frameworks require larger amounts of seed data to maintain a ground-truth language distribution~\cite{rosenbaum2022linguist,kim2021linda}, and are more costly than prompting PLMs~\cite{chen2022weakly}. However, in the context of dataset synthesis, seed data and label correctness are less important considerations. There is no task distribution from which seed data is drawn that PLMs must remain faithful to, and similarly, invariant ground-truth knowledge for language models is dependent on the desired task being synthesized.

Our work differs from existing applications of prompting for conversations along several dimensions.
Many studies examine utterance-level generation~\cite{chen2022weakly,sahu2022data,aher2022using,rosenbaum2022linguist}, whereas our work concerns the synthesis of full conversations.
\citet{bae-etal-2022-building} generated conversations for a narrow task and provided evaluations between their synthesis conditions. Recent concurrent work by \citet{kim2022soda} sought to distill conversations from InstructGPT 175B using a commonsense knowledge graph. In our work, we synthesize conversations using an open-source PLM and demonstrate that they are comparable to human-collected datasets, in terms of both conversation quality and usability as a dataset. Moreover, all of these studies only concern dyadic conversations, because the vast majority of conversational tasks are dyadic. Our work is the first study to synthesize multi-party conversations.
\section{Conversation Generation}
In this section, we discuss our methods for conversation generation. We first detail the construction of our example conversations, then describe their application to prompting PLMs.
\subsection{Writing Conversation Examples}
We simply wrote a pool of ten conversations between two speakers representing everyday dialogue using proper grammar. Along with each conversation, we wrote a brief conversation ``recipe'' which includes a topic, as well as  \textit{background information} for the two speakers\footnote{The first-author spent approximately 45 minutes on this writing process.}. 

The \textit{background information} represents some more fine-grained information about the two speakers, relevant to that particular topic. For example, Figure~\ref{fig:prompt_example} depicts an example prompt with three in-context conversation demonstrations. Each conversation is prefixed by a recipe and is structured in the same manner: ``The following is a conversation between Alice and Bob about \textit{topic}'' (e.g., ``pets'') followed by detailed background information (e.g., ``Alice love cats. Bob is more of a dog person.''). 

\subsection{Creating Conversations via Prompting}
Each prompt consists of three randomly sampled conversations from the aforementioned pool, along with their accompanying recipe.
After experimenting with PLMs of three different sizes (GPT-J 6B, GPT-NeoX 20B, OPT 30B), we primarily use OPT-30B and generate with nucleus
sampling with  $p=0.92$. Inspired by the format of DailyDialog, our handwritten and synthetically generated conversations fall into three categories: start-to-finish conversations, excerpts from the start to the middle of a conversation, and excerpts from the middle of a conversation. Several examples are given in the Appendix.

In this paper, we generate a dataset using a list of topics and tasks (i.e., subtopics) from the training set of the Feedback for Interactive Talk \& Search Dataset (FITS; \citet{xu2022learning}), a human-chatbot dataset designed to determine desirable human-chatbot tasks/conversations. FITS contains 5592 conversations which span 52 conversational topics (e.g., ``nutrition,'' ``philosophy'') with 315 subtopics (e.g., ``Italian food,'' ``Soren Kierkegaard''). We wrote background information for each of the 315 subtopics in the form given in Figure~\ref{fig:prompt_example}. 

Using the product of this process once results in a new synthetic dataset with 5592 conversations using the same topic, subtopic pairings from FITS. The average length of each conversation is 9.29 turns, with 12.84 words per turn. This is comparable to the dataset statistics of DailyDialog and Topical Chat, as per Table \ref{conv_stats}. In the Appendix, we have included the 315 prompt headers (Tables \ref{DyadicPrompts}, \ref{TriadicPrompts}) and the pool of in-context examples (Tables \ref{HWExamples}, \ref{HWTriadicExamples}, \ref{HWTriadicExamplesPt2}). 

\vspace{-2mm}
\section{Synthetic Conversation Evaluation}
\label{conversation_evaluation}

In Figure~\ref{fig:Examples}, the top-left is taken from DailyDialog, whereas the top-right is generated synthetically. The bottom-left is generated synthetically and the bottom-right is taken from MPC. 
\begin{table}[]
\centering
\small
\begin{tabular}{lcc}
\toprule
Source & Words/Turn & Turns/Conv.\\ 
\midrule
DailyDialog & 11.58 & ~7.84 \\
Topical Chat & 13.38 & 21.83 \\
[5pt]
HW Examples & 11.00 & ~8.10 \\
Synthetic & 10.70 & ~9.29 \\
\bottomrule
\end{tabular}
\caption{Number of words per turn and number of turns per conversation for all conversations. HW Examples represents the ten handwritten conversation examples, and Synthetic represents synthetic conversations generated using OPT 30B.}
\label{conv_stats}
\vspace{-4mm}
\end{table}


\subsection{Evaluation of Conversation Quality}
\label{sec:conversation_quality_evaluation}
Table~\ref{exp_1a} provides a crowdworker evaluation of our synthetic dataset compared against DailyDialog and Topical Chat. We expect Topical Chat to be rated as the most interesting, due to the knowledge-grounding process utilized during the dialogue collection process. We randomly sampled 200 conversations for each conversation source and asked a pre-qualified pool of 28 crowdworkers on Amazon Mechanical Turk (AMT) to rate each conversation. 
The instructions and details of our human evaluation setup are explained in Appendix~\ref{sec:HumanEvaluation}.

\begin{table}[t]
\centering
\small
\setlength{\tabcolsep}{3pt}
\scalebox{0.99}{
\begin{tabular}{lccccc}
\toprule
Source        & Interesting & Coherent & Natural & Consistent 
\\ 
\midrule
DailyDialog & 3.44 & 4.51 & 4.85 & 4.57 \\
Topical Chat & \textbf{4.55} & 4.39 & \textbf{4.92} & \textbf{4.87}  \\
[5pt]
GPT-J 6B & 3.96${}^{*}$ & 4.49 & 4.86 & 4.36 \\
GPT-NeoX 20B & 3.81${}^{*}$ & 4.40 & 4.63 & 4.35 \\
OPT 30B & 4.13${}^{*}$ & \textbf{4.61${}^{*\dagger}$} & 4.82 & 4.63  \\
\bottomrule
\end{tabular}
}
\caption{Evaluation of conversations randomly sampled from DailyDialog, Topical Chat, and three synthetic datasets generated by prompting GPT-J 6B, GPT-NeoX 20B, and OPT 30B. $^*$ indicates statistical significance over DailyDialog. ${}^\dagger$ indicates statistical significance over Topical Chat. Significance computed at $\alpha=0.05$.}
\label{exp_1a}
\end{table}

As these conversations are generated using prompting, we first checked whether each conversation followed the prescribed prompt. Crowdworkers identified 95\% of the conversations generated by OPT 30B as matching the topic stated in the prompt\footnote{91\% and 92\% for GPT-J 6B and GPT-NeoX 20B.}, indicating this prompting strategy's effectiveness for topic-grounded conversation generation. Overall, Table~\ref{exp_1a} indicates that synthetic conversations generated by OPT 30B are rated as the most coherent, and more interesting and consistent than DailyDialog. The synthetic conversations are almost as natural as DailyDialog, but are rated as less interesting and natural than Topical Chat. Given our results, we also hypothesize that larger models likely produce higher quality conversations. We provide several examples of conversations generated by OPT 175B using an online web interface\footnote{https://opt.alpa.ai/} in the Appendix.

\begin{figure}[t]
    \centering
    \includegraphics[width=\linewidth]{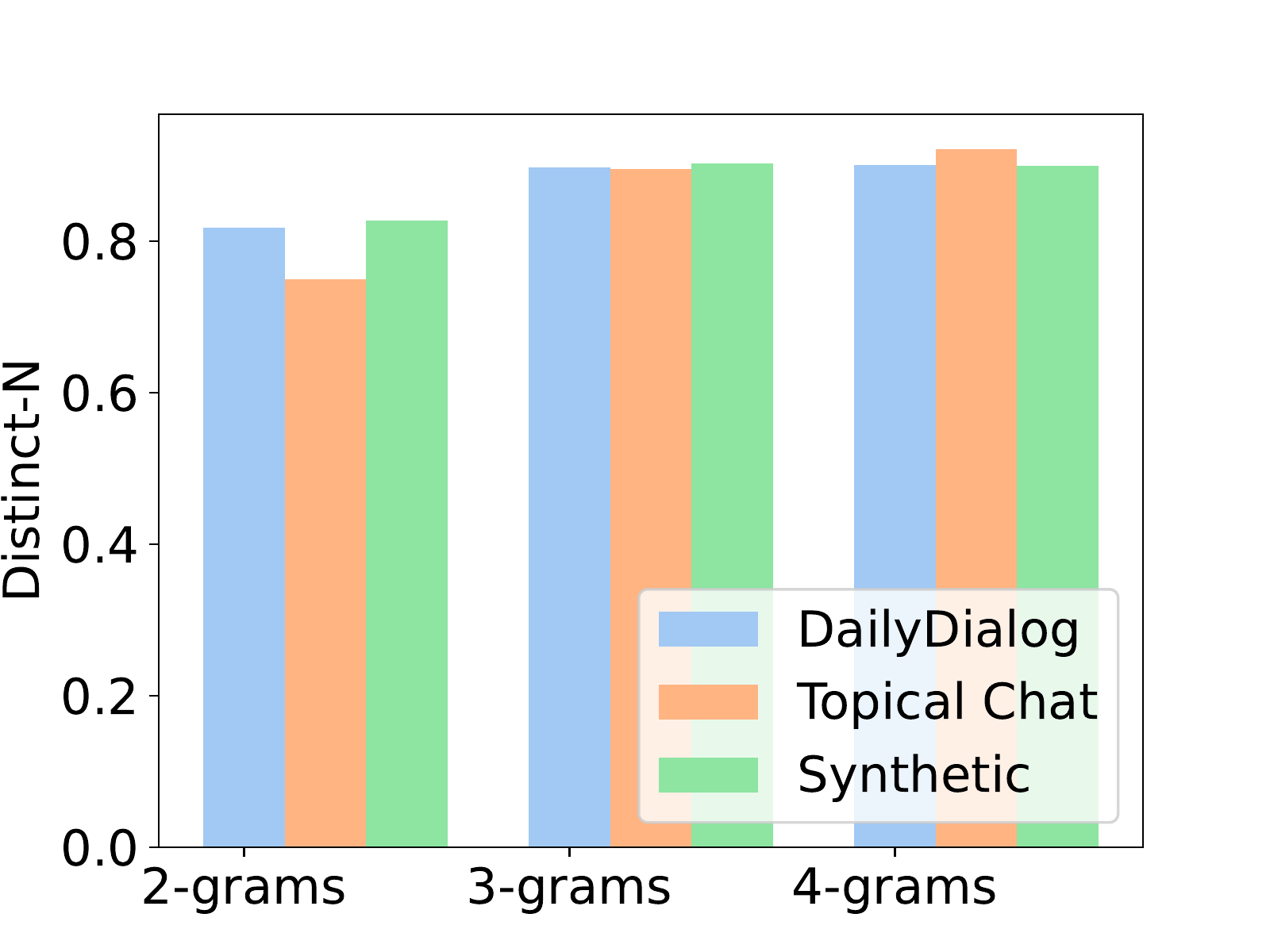}
    \caption{Distinct-N with $N=2,3,4$ for conversations in DailyDialog, Topical Chat, and our synthetic conversations. Our synthetic conversations have the highest most unique bi-grams and tri-grams, and the second-most unique 4-grams.}
    \label{fig:exp1_diversity}
    \vspace{-4mm}
\end{figure}

A concern one might have is that since in-context examples heavily influence prompting~\cite{min2022rethinking,lu2021fantastically}, our small in-context example size may limit the lexical diversity of our synthetic conversations. Following earlier work evaluating text generation, we use Distinct-N to measure lexical diversity \cite{wu2021textgail, li2016diversity}. Figure~\ref{fig:exp1_diversity} shows that our synthetically generated conversations are slightly more diverse than both DailyDialog and Topical Chat in terms of distinct bigrams and trigrams, and slightly less diverse than Topical Chat in terms of 4-grams.

We then sought to examine the impact of using expert handwritten examples by comparing against synthetic conversations generated using conversations from DailyDialog and Topical Chat as in-context examples. We set the number of conversation examples such that the number of in-context dialogue turns are approximately equal across all conditions. Table~\ref{exp_1b} shows that synthetic conversations generated conditioned on handwritten in-context examples are the most coherent, natural, and on-topic. In terms of interestingness and consistency, the ratings of these conversations slightly trail the ratings of the conversations generated conditioned on Topical Chat.

\begin{table}[]
\centering
\small
\begin{tabular}{llll}
\toprule
Dimension & DD-IC & TC-IC & HW-IC \\ 
\midrule
Interesting & 3.82 & \textbf{4.35} & 4.27$^*$ \\
Coherent & 4.48 & 4.56 & \textbf{4.77}$^{*+}$ \\
Natural & 4.54 & 4.69 & \textbf{4.69}$^*$ \\
Consistent & 4.76 & \textbf{4.87} & 4.86$^*$ \\
On-Topic & 0.91 & 0.88 & \textbf{0.96}$^{*+}$\\
\bottomrule
\end{tabular}
\caption{Human evaluation of conversations generated using OPT-30B with in-context examples randomly sampled from DailyDialog (DD-IC), Topical Chat (TC-IC), and handwritten examples (HW-IC). $^*$ indicates statistical significance over DD-IC and $^+$ indicates statistical significance over TC-IC.}
\label{exp_1b}
\vspace{-4mm}
\end{table}
\subsection{Fine-Tuning with Synthetic Conversations}
After establishing that our synthetic conversations are of rather high quality on their own, we attempted to use the synthetic dataset as training data for dialogue models. We fine-tuned distilled BlenderBot 400M~\cite{roller-etal-2021-recipes} on DailyDialog, Topical Chat, and our synthetic conversations\footnote{For fair comparison, we fine-tune on the same numebr of training instances via downsampling.}. 

Rather than directly prompting OPT as a response generator, we select BlenderBot as a lightweight, effective dialogue model. This allows for comparisons between the three data sources as training sets, because fine-tuning OPT is prohibitively expensive. Moreover, while prompting with larger PLMs can yield coherent responses, it is generally impractical as an end-to-end dialogue system if hosted on typically available hardware. For long inputs (e.g. with multiple dialogues in-context), generation time typically takes several minutes using OPT 30B\footnote{All experiments are conducted using one p3dn.24xlarge AWS EC2 instance.}.

We first performed an interactive human evaluation of the three dialogue models as end-to-end social chatbots using the LegoEval platform~\cite{li2021legoeval}. 
Details can be found in Appendix~\ref{sec:HumanEvaluation}.

\begin{table}[]
\centering
\small
\begin{tabular}{llll}
\toprule
Dimension & DD & TC & Syn \\ 
\midrule
Interesting & 3.35 & \textbf{3.86} & 3.30 \\
Coherent & 3.52 & \textbf{3.71} & 3.68 \\
Natural & 3.52 & 3.57 & \textbf{3.68} \\
Consistent & 3.35 & \textbf{3.65} & 3.32 \\
Engaging & 3.73 & \textbf{3.88} & 3.65 \\
Intelligent & 3.41 & \textbf{3.55} & 3.24 \\
Non-repetitive & 3.37 & 3.37 & \textbf{3.40} \\
\bottomrule
\end{tabular}
\caption{Interactive human evaluation yields comparable ratings for chatbots fine-tuned on conversations from DailyDialog (DD), Topical Chat (TC), and our Synthetic Data (Syn). 
}
\label{exp_2}
\vspace{-4mm}
\end{table}
Table~\ref{exp_2} shows that dialogue models fine-tuned on our synthetic conversations are rated comparably to dialogue models fine-tuned on real human-human data --- the chatbot fine-tuned on synthetic data appeared to be the most natural and non-repetitive, and was rated as the second-most coherent. It was rated as the least intelligent, engaging, consistent, and interesting. However, two-sided t-tests at $\alpha=0.05$ revealed that there was not a statistically significant difference in ratings between the models fine-tuned on all three datasets across all dimensions except for interestingness. The Topical Chat model was rated as significantly more interesting, as expected.

In terms of automatic evaluation, we applied these dialogue models on out-of-distribution test sets to prevent an unfair comparison. We evaluated models fine-tuned on DailyDialog and our synthetic data on Topical Chat, and models fine-tuned on Topical Chat and our synthetic data on DailyDialog. Table~\ref{tab:exp_2_automatic} indicates that in terms of perplexity and ROUGE, models fine-tuned on our synthetic data generalize to out-of-distribution convesational data as well as models trained on real human-human datasets. On the DailyDialog test set, the synthetic dataset model outperforms the Topical Chat model on all metrics except ROUGE-2, and on the Topical Chat test set, the synthetic dataset model underperforms the DailyDialog model on all metrics except perplexity.




\section{Triadic and Multi-Party Conversations}
\begin{table}[]
    \centering
    \small
    \begin{tabular}{lccc}
        \toprule
        Metric (Test Set) & DD-BB & TC-BB & Syn-BB \\
        \midrule
        Perplexity (DD) & --- & 120.2 & \textbf{87.05}\\
        ROUGE-1 (DD) & --- & 12.34 & \textbf{12.90}\\
        ROUGE-2 (DD) & --- & \textbf{~1.66} & ~1.52\\
        ROUGE-L (DD) & --- & 10.60 & \textbf{10.94}\\
        [5pt]
        Perplexity (TC) & ~43.3 & --- & \textbf{37.1} \\
        ROUGE-1 (TC) & \textbf{16.63} & --- & 15.13 \\
        ROUGE-2 (TC) & \textbf{~2.36} & --- & ~1.77\\
        ROUGE-L (TC) & \textbf{13.61} & --- & 12.41 \\
        \bottomrule
    \end{tabular}
    \caption{Out-of-distribution automatic evaluation of perplexity and ROUGE is comparable for BlenderBot fine-tuned on DailyDialog (DD-BB), Topical Chat (TC-BB), and synthetic data generated using our handwritten examples in-context (Syn-BB), respectively.}
    \label{tab:exp_2_automatic}
    \vspace{-4mm}
\end{table}
The vast majority of dialogue tasks and conversational datasets focus on dyadic conversations (e.g. \citet{li2017dailydialog,gopalakrishnan2019topical,smith2020can,rashkin2019towards}), following the traditional speaker-listener paradigm~\cite{engelhardt2006speakers}. In contrast, the literature on multi-party social conversation is rather scarce, not only in terms of conversation generation but as a task altogether. However, while it is an understudied research area, it is incredibly important, because dyadic conversations do not capture the full reality of in-person, human-human social conversations, nor the full potential of dialogue agents. To name a few applications, dialogue agents have the potential to supplement classroom learning with multiple parties, serving as a third mediating party in a debate or discussion between two people, or to provide companionship and support in virtual group settings. A major reason why these lines of work remain unsolved is that there are few large-scale multi-party dialogue datasets. 

Many existing multi-party datasets are scripted corpora such as MELD~\cite{poria2019meld} or MPDD~\cite{chen2020mpdd} or HLA-Chat~\cite{ju-etal-2022-learning,li2020aloha}. Other multi-party corpora are collected for highly domain-specific purposes, such as multi-party empathetic dialogue~\cite{zhu-etal-2022-multi}. Such corpora are also typically collected through asynchronous online platforms, rather than natural conversation. These platforms exist in the form of forums and online chat platforms such as Ubuntu IRC~\citep{lowe2015ubuntu} or Reddit~\citep{baumgartner2020pushshift}. Other more natural multi-party conversational datasets are license-protected speech datasets (e.g. CHIME~\citep{christensen2010chime}) which have been constructed for tasks such as speaker attribution. 

\begin{figure}[t]
    \centering
    \includegraphics[width=\linewidth]{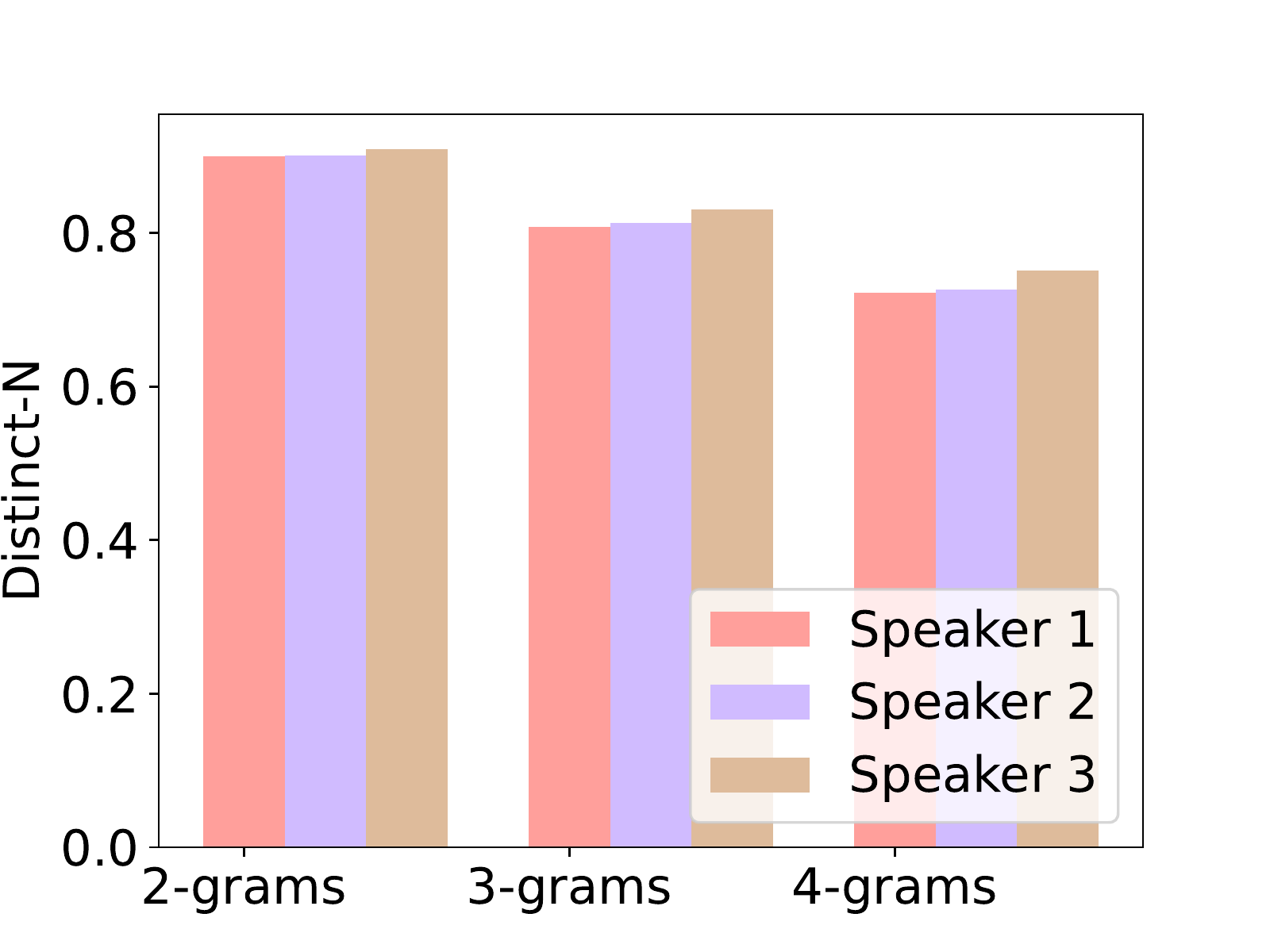}    
    \caption{Linguistic diveristy (Distinct-N) is comparable for each speaker in the synthetic triadic conversation dataset.}
    \label{fig:triadic_diversity_by_speaker}
    \vspace{-4mm}
\end{figure}
We find that we can apply our prompting approach to generate synthetic, open-domain, multi-party social conversations following the same structure as our synthetic dyadic conversations\footnote{While we effectively use Alice, Bob, and Claire instead of Speaker 1, Speaker 2, and Speaker 3, respectively, the order of speakers does not necessarily follow the speaker order in the in-context examples (e.g. Appendix Table~\ref{TriadicExample1}).}. As in the dyadic case, we generate triadic conversations using optional background information for each speaker. We consider the ``Multi-Party Chat'' corpus (MPC) \cite{shaikh2010mpc}, a text-based, open-domain conversation dataset collected in real-time online sessions at the University of Albany, and MELD, which contains scripted multi-party dialogues from the popular sitcom ``Friends.'' We directly compare our synthetically generated conversations against MPC and MELD.

Table~\ref{exp_4} includes our evaluation of our conversations using the same pool of pre-qualified AMT workers, again with 200 randomly sampled conversations. MPC consists of massive conversation settings --- on the scale of 500 turns for a typical conversation session --- so we randomly sample 8 to 12\footnote{The length between 8 and 12 turns is chosen uniformly.} continuous turns for each conversation evaluation to more closely match the structure of our synthetic conversations.\footnote{We sample rather than selecting the first 8-12 turns, to avoid overrepresenting greetings.} We present examples of MPC  and MELD in Appendix Tables~\ref{MPCExample1}, \ref{MELDExample}. 

We inform the AMT workers that they will read conversation excerpts. In addition to the questions in Table~\ref{exp_1a}, we add two questions specific to multi-party conversations. We ask if the conversation excerpt looks comprehensible (in terms of the reader being able to determine who each speaker is addressing), and we ask if all parties of the conversation are participating equally and actively.

In Table~\ref{exp_4}, we find that the synthetic conversations are rated statistically significantly more favorably than MPC and MELD across all dimensions. Beyond conversation quality, it is possible that the ratings for MPC are comparatively low due to the fact that each conversation typically has more than three speakers, which may be more difficult for human raters to interpret. Our results for MELD also indicate that while the corpus is high quality, it may be better fit for comedy and accompaniment with visual context, than as pure dialogue.

Additionally, we checked the linguistic diversity for each speaker. In terms of Distinct-N, each speaker's lexical diversity is comparable (Figure~\ref{fig:triadic_diversity_by_speaker}) as well as the number of words per turn (12.2, 12.2, and 13.5 for Speakers 1, 2, and 3 respectively). The triadic conversations tended to be slightly longer than the average dyadic conversation (11.5 turns/conversation versus 9.29 turns/conversation). 
\begin{table}[]
\centering
\small
\begin{tabular}{lccc}
\toprule
Dimension & MPC & MELD & Syn \\ 
\midrule
Interesting & 2.48 & 3.52 & \textbf{4.14}$^*$ \\
Coherent & 2.40 & 3.68 & \textbf{4.65}$^*$ \\
Natural & 2.69 & 3.69 & \textbf{4.47}$^*$ \\
Consistent & 2.96 & 3.83 & \textbf{4.65}$^*$ \\
Comprehensible & 2.48 & 3.83 & \textbf{4.80}$^*$ \\
Balanced Engagement & 3.45 & 4.00 & \textbf{4.89}$^*$ \\
\bottomrule
\end{tabular}
\caption{Synthetic conversations generated using OPT 30B are rated significantly higher than MPC and MELD across all dimensions. 
}
\label{exp_4}
\vspace{-4mm}
\end{table}

\vspace{-4mm}
\section{Discussion}
Overall, we find that prompting PLMs to generate synthetic conversations is promising. 
\subsection{Considerations for Dyadic Dialogue}
The synthetically generated conversations appear comparable to conversations from human-collected datasets. The individual conversations appear interesting, coherent, natural, and consistent, as the average ratings for each category lie between 4.0 and 5.0. The Appendix includes multiple examples of conversations generated using the strongest performing PLM (OPT 30B, e.g. Table~\ref{DyadicExample1}) as well as several conversations generated using OPT 175B (e.g. Table~\ref{DyadicExample2}). Tables~\ref{exp_2} and \ref{tab:exp_2_automatic} also indicate that fine-tuning on synthetically generated examples can result in dialogue models of comparable quality, with the potential for further improvements by simply generating more synthetic conversations.

Future work may consider applying applying this generation approach to dyadic contexts beyond social conversations, such as task-oriented dialogue. The clearest difference between social and task-oriented dialogue contexts is the importance of knowledge grounding. In task-oriented dialogue, there typically needs to be retrieval from knowledge base for response generation. An application of PLACES could involve using database results as a ground-truth reference. Rather than using a topic list like FITS, one could form conversational recipes using database search results as background information. Given the apparent semantic control described in Section~\ref{conversation_evaluation}, it is possible that synthetic task-oriented conversations would be able to correctly utilize knowledge.

\subsection{Considerations for Multi-Party Dialogue}
\label{sec:multiparty_considerations}
We found that in comparison to MPC, our synthetic triadic dialogues appear to be of fairly high quality. However, there remain several open questions about multi-party dialogue, even in the triadic case. For instance, there is not a set archetype of conversations. Sometimes, conversations may be dominated by a single speaker, whereas in others, each speaker in the conversation may contribute equally. Depending on the scenario, a speaker may be the facilitator --- meetings can be considered (topic-specific) multi-party dialogues which are typically led by designated speakers.

Moreover, there are several questions about how to utilize multi-party dialogues in an interactive dialogue system. There are use cases where it may be appropriate for one dialogue system to interact with multiple users. On the other hand, in scenarios like emotional support dialogue systems, it may make sense for a single user to interact with multiple simulated conversational parties.

Here, we investigated our approach's potential to generate synthetic multi-party conversations, hoping to bridge the gap in data availability in multi-party chat. This opens opportunities for a variety of applications. Synthetic datasets could be used to help discover how to properly model triadic and multi-party conversations. In the future, datasets could also be generated for domain-specific, multi-party applications ranging from language learning to task-oriented spoken dialogue systems.

\section{Error Analysis}
We examine the dyadic and triadic conversations which received low scores (1/5) across multiple dimensions.

\vspace{-2mm}
\subsection{Dyadic Conversations}
Out of the dyadic conversations, two conversations were rated as generic and dull. One conversation (Appendix Table~\ref{ErrorAnalysis1}) talks about the singer, Taylor Swift. However, the conversation is repetitive, repeating utterances such as ``What are your thoughts on her?'' and ``I think she is very nice.'' The other conversation is about the filmmaker, Ken Burns (Appendix Table~\ref{ErrorAnalysis2}). While the conversation is appears coherent and uses correct factual information (e.g., making reference to Ken Burns' documentaries on World War II and the Vietnam War), the language could be perceived as dull. 

Three conversations were rated as completely unnatural. In one case, the PLM missed the prescribed subtopic (cotton candy) and instead hallucinated a conversation about a sensitive topic, cancer (Appendix Table~\ref{ErrorAnalysis4}). This is also the only conversation to be rated as completely incoherent. The other two conversations are both on-topic. However, one conversation is on-topic but rather short (five turns), whereas the other conversation is overly verbose and a little repetitive.

There were also three conversations were evaluated as completely inconsistent. In all three conversations, the roles of the two speakers seemingly swap.
While these hypothetical turns are possible in excerpts of real conversations, they assume background information or events which have not been explicitly established when considered as standalone conversations. An example is given in Appendix Table~\ref{ErrorAnalysis8}.

While some of the evaluations may be subjective, an issue that has objectively appeared multiple times is the consistency of speakers' utterances. The intents and personas of the speakers appear to get switched, which is also an open problem in dialogue systems research. Future work may look to combine conversation synthesis approaches with strategies for dialogue consistency such as the generate-delete-rewrite framework~\cite{song-etal-2020-generate} or language inference approaches~\cite{welleck-etal-2019-dialogue,song2020generating}.
\vspace{-2mm}

\subsection{Triadic Conversations}
No conversations were perceived as completely incomprehensible, but human evaluators indicated that two conversations appeared to have imbalanced engagement --- in both cases, the third speaker (``Claire'') only has one dialogue turn. As discussed in Section~\ref{sec:multiparty_considerations}, however, it is not clear whether this is a drawback. Real-life triadic conversations do not follow a set archetype in terms of engagement balance.

There was one conversation which was rated as completely incoherent. In the conversation, there is one dialogue turn which presents information inconsistent with prior turns, but the another issue appears to be an oddly placed transition which brings the conversation from travel to hobbies: ``You should definitely go to Paris! What do you like to do for fun?'' (Appendix Table~\ref{ErrorAnalysis9}).

There are two conversations which were perceived as completely unnatural. However, naturalness appears to be a rather subjective evaluation. One conversation is given in Appendix Table~\ref{ErrorAnalysis10}, and it is debatable whether the language conventions used are unnatural. One could argue that it is overly enthusiastic, but others could argue that it is how some people speak colloquially. Interestingly, the second conversation which received a low naturalness score is also enthusiastic and about the same topic (gardening).

The only conversation which was rated as generic and dull was a 15-turn debate about whether the European Union is a ``conspiracy'' (Appendix Table~\ref{ErrorAnalysis12}). The debate is rather shallow and does not make a lot of progress.

As with the dyadic conversation error analysis, we see that there are issues with persona consistency. However, unlike the dyadic scenario, there are fewer existing solutions for dialogue consistency. Multi-party conversation synthesis could potentially be improved by applying ideas from the newly published PersonaTKG dialogue system, which employs a unified graph that encodes personas, utterances, and external knowledge on a scripted dialogue dataset~\cite{ju-etal-2022-learning}. 

Beyond consistency, in the example from Table~\ref{ErrorAnalysis12} we see that there is potential for PLMs to hallucinate misinformation. There are again fewer existing studies on circumventing this obstacle in multi-party dialogue, but future work could look to incorporating external knowledge~\cite{kang2022knowledge} or dialogue safety approaches~\cite{kim2021robust,dinan-etal-2019-build}. All said, our work motivates further study into multi-party dialogue consistency, safety, and synthesis.

\section{Conclusion}
In this work, we presented an application of prompting PLMs to create synthetic conversations. These synthetic conversations are comparable in terms of quality and lexical diversity to actual human-human datasets, and can be used as training data for dialogue models. This opens avenues in generative language work such as collaborative and creative writing, story generation, as well as synthesis of new conversational tasks. Here, we presented one example --- synthesizing a multi-party conversational dataset. This presents a unique opportunity to further study multi-party dialogue modeling.

\section{Limitations}
\paragraph{Controllability.} We witness encouraging levels of control through the prompt ($95\%$ of the time, the synthetic conversation matches the desired topic), but prompting PLMs is still an uncontrolled form of generation. Future work could seek to add more semantic controls beyond the stated topic in the prompt or explore using weak supervision to provide post-hoc improvements on synthetic data quality, similar to \citet{chen2022weakly}. In this work, we also did not thoroughly explore the effects of different generation approaches. Future work may consider applying semantic constraints during the decoding process~\cite{lu-etal-2021-neurologic}. Further controls are necessary before using this approach for higher-stakes settings such as task-oriented dialogue and other knowledge-grounded tasks.

\paragraph{Cost of Human Effort.} While we demonstrate the ability to synthesize large amounts of data, the quality of a synthesized dataset is still dependent on human effort, to an extent. One can use a generic prompt template such as ``Alice is interested in [subtopic]'' for each subtopic, but we qualitatively see that more detailed background information in a prompt often yields better generation performance.

In this work, we generated 5592 dyadic and triadic conversations, matching the number of topic combinations in FITS. PLACES can be used to generate many more conversations in the future. Using the same overall can continue to make new combinations of topic and subtopic, or simply rerun the generation process as it is nondeterministic. Moreover, one may consider filling the slots in our conversation recipes using an abundant of external sources, including from existing dataset annotations (e.g. Persona Chat~\citet{zhang2018personalizing}).

\paragraph{Computational Costs.} Once a dataset is synthesized, small, task-specific models can be used downstream. However, the synthesis method used in this work is still expensive: we prompt PLMs. While we only used freely accessible PLMs such as OPT, we acknowledge that not everyone has access to the number of GPUs necessary to load PLMs, even for inference.

\paragraph{Prompt Design.} The idea of prompting large language models is not novel. There is a plethora of work that examines how to apply prompting to a variety of different tasks (e.g. \citet{brown2020language,min2021metaicl}), along with several studies on how to mine or engineer different prompts~\cite{liu2021pre}. In this work, we do not claim novelty to our prompt, nor do we claim that our prompt design is the optimal prompt for conversation generation. Our prompt is designed in a conversational manner, drawing inspiration from \citet{chen2022weakly}. We instead emphasize the application of prompting for conversational dataset synthesis. The idea of synthesizing conversational datasets ``from scratch'' is previously unexplored, and has potential to supplement a lot of areas of dialogue research, such as multi-party conversations.

\section{Ethical Considerations}
\paragraph{Human Evaluation and Crowdsourcing.} We make use of crowdsourcing through Amazon Mechanical Turk for several experiments. All crowdworkers were paid at a rate higher than the minimum wage in California. In accordance with California State Law, all crowdworkers were also informed they were speaking with chatbots during the data collection for our interactive evaluation. All participants consented to the logging of their responses.

\paragraph{Language Model Biases.} Large pre-trained language models are typically pre-trained on massive corpora crawled from the internet such as The Pile~\cite{gao2020pile} or Common Crawl. This allows language models to have exposure to a large amount of linguistic diversity, but this also results in exposure to a lot of hateful, biased, or otherwise undesirable content from the internet~\cite{luccioni2021s}. Future work should examine combining conversation synthesis with dialogue safety approaches.

\paragraph{Scientific Artifacts.} 
All scientific artifacts are used according to their intended purpose. The FITS dataset is publicly available at \url{https://parl.ai/projects/fits/}. OPT is an open-source language model. GPT-J is available for use under the MIT license. We use the HuggingFace Transformers and PyTorch packages for all modeling~\cite{wolf2020transformers, paszke2019pytorch}. All artifacts used are in English.

\bibliographystyle{acl_natbib}
\bibliography{references}

\begin{thebibliography}{63}
\expandafter\ifx\csname natexlab\endcsname\relax\def\natexlab#1{#1}\fi

\bibitem[{Aher et~al.(2022)Aher, Arriaga, and Kalai}]{aher2022using}
Gati Aher, Rosa~I Arriaga, and Adam~Tauman Kalai. 2022.
\newblock Using large language models to simulate multiple humans.
\newblock \emph{arXiv preprint arXiv:2208.10264}.

\bibitem[{Allahbakhsh et~al.(2013)Allahbakhsh, Benatallah, Ignjatovic,
  Motahari-Nezhad, Bertino, and Dustdar}]{allanbaksh2013quality}
Mohammad Allahbakhsh, Boualem Benatallah, Aleksandar Ignjatovic, Hamid~Reza
  Motahari-Nezhad, Elisa Bertino, and Schahram Dustdar. 2013.
\newblock \href {https://doi.org/10.1109/MIC.2013.20} {Quality control in
  crowdsourcing systems: Issues and directions}.
\newblock \emph{IEEE Internet Computing}, 17(2):76--81.

\bibitem[{Bae et~al.(2022)Bae, Kwak, Kim, Ham, Kang, Lee, and
  Park}]{bae-etal-2022-building}
Sanghwan Bae, Donghyun Kwak, Sungdong Kim, Donghoon Ham, Soyoung Kang, Sang-Woo
  Lee, and Woomyoung Park. 2022.
\newblock \href {https://doi.org/10.18653/v1/2022.naacl-main.155} {Building a
  role specified open-domain dialogue system leveraging large-scale language
  models}.
\newblock In \emph{Proceedings of the 2022 Conference of the North American
  Chapter of the Association for Computational Linguistics: Human Language
  Technologies}, pages 2128--2150, Seattle, United States. Association for
  Computational Linguistics.

\bibitem[{Baumgartner et~al.(2020)Baumgartner, Zannettou, Keegan, Squire, and
  Blackburn}]{baumgartner2020pushshift}
Jason Baumgartner, Savvas Zannettou, Brian Keegan, Megan Squire, and Jeremy
  Blackburn. 2020.
\newblock The pushshift reddit dataset.
\newblock In \emph{Proceedings of the international AAAI conference on web and
  social media}, volume~14, pages 830--839.

\bibitem[{Brown et~al.(2020)Brown, Mann, Ryder, Subbiah, Kaplan, Dhariwal,
  Neelakantan, Shyam, Sastry, Askell et~al.}]{brown2020language}
Tom Brown, Benjamin Mann, Nick Ryder, Melanie Subbiah, Jared~D Kaplan, Prafulla
  Dhariwal, Arvind Neelakantan, Pranav Shyam, Girish Sastry, Amanda Askell,
  et~al. 2020.
\newblock Language models are few-shot learners.
\newblock \emph{Advances in neural information processing systems},
  33:1877--1901.

\bibitem[{Chen et~al.(2022)Chen, Papangelis, Tao, Rosenbaum, Kim, Liu, Yu, and
  Hakkani-Tur}]{chen2022weakly}
Maximillian Chen, Alexandros Papangelis, Chenyang Tao, Andy Rosenbaum, Seokhwan
  Kim, Yang Liu, Zhou Yu, and Dilek Hakkani-Tur. 2022.
\newblock Weakly supervised data augmentation through prompting for dialogue
  understanding.
\newblock NeurIPS 2022 Workshop on Synthetic Data for Empowering ML Research.

\bibitem[{Chen et~al.(2020)Chen, Huang, and Chen}]{chen2020mpdd}
Yi-Ting Chen, Hen-Hsen Huang, and Hsin-Hsi Chen. 2020.
\newblock Mpdd: A multi-party dialogue dataset for analysis of emotions and
  interpersonal relationships.
\newblock In \emph{Proceedings of the 12th Language Resources and Evaluation
  Conference}, pages 610--614.

\bibitem[{Christensen et~al.(2010)Christensen, Barker, Ma, and
  Green}]{christensen2010chime}
Heidi Christensen, Jon Barker, Ning Ma, and Phil~D Green. 2010.
\newblock The chime corpus: a resource and a challenge for computational
  hearing in multisource environments.
\newblock In \emph{Eleventh Annual Conference of the International Speech
  Communication Association}. Citeseer.

\bibitem[{Dinan et~al.(2019)Dinan, Humeau, Chintagunta, and
  Weston}]{dinan-etal-2019-build}
Emily Dinan, Samuel Humeau, Bharath Chintagunta, and Jason Weston. 2019.
\newblock \href {https://doi.org/10.18653/v1/D19-1461} {Build it break it fix
  it for dialogue safety: Robustness from adversarial human attack}.
\newblock In \emph{Proceedings of the 2019 Conference on Empirical Methods in
  Natural Language Processing and the 9th International Joint Conference on
  Natural Language Processing (EMNLP-IJCNLP)}, pages 4537--4546, Hong Kong,
  China. Association for Computational Linguistics.

\bibitem[{Engelhardt et~al.(2006)Engelhardt, Bailey, and
  Ferreira}]{engelhardt2006speakers}
Paul~E Engelhardt, Karl~GD Bailey, and Fernanda Ferreira. 2006.
\newblock Do speakers and listeners observe the gricean maxim of quantity?
\newblock \emph{Journal of memory and language}, 54(4):554--573.

\bibitem[{Gao et~al.(2020)Gao, Biderman, Black, Golding, Hoppe, Foster, Phang,
  He, Thite, Nabeshima et~al.}]{gao2020pile}
Leo Gao, Stella Biderman, Sid Black, Laurence Golding, Travis Hoppe, Charles
  Foster, Jason Phang, Horace He, Anish Thite, Noa Nabeshima, et~al. 2020.
\newblock The pile: An 800gb dataset of diverse text for language modeling.
\newblock \emph{arXiv preprint arXiv:2101.00027}.

\bibitem[{Gopalakrishnan et~al.(2019)Gopalakrishnan, Hedayatnia, Chen,
  Gottardi, Kwatra, Venkatesh, Gabriel, Hakkani-T{\"u}r, and
  AI}]{gopalakrishnan2019topical}
Karthik Gopalakrishnan, Behnam Hedayatnia, Qinglang Chen, Anna Gottardi,
  Sanjeev Kwatra, Anu Venkatesh, Raefer Gabriel, Dilek Hakkani-T{\"u}r, and
  Amazon~Alexa AI. 2019.
\newblock Topical-chat: Towards knowledge-grounded open-domain conversations.
\newblock In \emph{INTERSPEECH}, pages 1891--1895.

\bibitem[{Ju et~al.(2022)Ju, Feng, Lv, Wang, and Zhang}]{ju-etal-2022-learning}
Dongshi Ju, Shi Feng, Pengcheng Lv, Daling Wang, and Yifei Zhang. 2022.
\newblock \href {https://aclanthology.org/2022.coling-1.23} {Learning to
  improve persona consistency in multi-party dialogue generation via text
  knowledge enhancement}.
\newblock In \emph{Proceedings of the 29th International Conference on
  Computational Linguistics}, pages 298--309, Gyeongju, Republic of Korea.
  International Committee on Computational Linguistics.

\bibitem[{Kang et~al.(2022)Kang, Kwak, Baek, and Hwang}]{kang2022knowledge}
Minki Kang, Jin~Myung Kwak, Jinheon Baek, and Sung~Ju Hwang. 2022.
\newblock Knowledge-consistent dialogue generation with knowledge graphs.
\newblock In \emph{ICML 2022 Workshop on Knowledge Retrieval and Language
  Models}.

\bibitem[{Kim et~al.(2021{\natexlab{a}})Kim, Kim, Hong, and
  Kim}]{kim2021robust}
Byeongchang Kim, Hyunwoo Kim, Seokhee Hong, and Gunhee Kim. 2021{\natexlab{a}}.
\newblock How robust are fact checking systems on colloquial claims?
\newblock In \emph{Proceedings of the 2021 Conference of the North American
  Chapter of the Association for Computational Linguistics: Human Language
  Technologies}, pages 1535--1548.

\bibitem[{Kim et~al.(2022)Kim, Hessel, Jiang, Lu, Yu, Zhou, Bras, Alikhani,
  Kim, Sap et~al.}]{kim2022soda}
Hyunwoo Kim, Jack Hessel, Liwei Jiang, Ximing Lu, Youngjae Yu, Pei Zhou,
  Ronan~Le Bras, Malihe Alikhani, Gunhee Kim, Maarten Sap, et~al. 2022.
\newblock Soda: Million-scale dialogue distillation with social commonsense
  contextualization.
\newblock \emph{arXiv preprint arXiv:2212.10465}.

\bibitem[{Kim et~al.(2021{\natexlab{b}})Kim, Jeong, and Cho}]{kim2021linda}
Yekyung Kim, Seohyeong Jeong, and Kyunghyun Cho. 2021{\natexlab{b}}.
\newblock Linda: Unsupervised learning to interpolate in natural language
  processing.
\newblock \emph{arXiv preprint arXiv:2112.13969}.

\bibitem[{Kulh{\'a}nek et~al.(2021)Kulh{\'a}nek, Hude{\v{c}}ek, Nekvinda, and
  Du{\v{s}}ek}]{kulhanek2021augpt}
Jon{\'a}{\v{s}} Kulh{\'a}nek, Vojt{\v{e}}ch Hude{\v{c}}ek, Tom{\'a}{\v{s}}
  Nekvinda, and Ond{\v{r}}ej Du{\v{s}}ek. 2021.
\newblock Augpt: Auxiliary tasks and data augmentation for end-to-end dialogue
  with pre-trained language models.
\newblock In \emph{Proceedings of the 3rd Workshop on Natural Language
  Processing for Conversational AI}, pages 198--210.

\bibitem[{Li et~al.(2020)Li, Jiang, Feng, Sprague, Zhou, and
  Hoey}]{li2020aloha}
Aaron~W Li, Veronica Jiang, Steven~Y Feng, Julia Sprague, Wei Zhou, and Jesse
  Hoey. 2020.
\newblock Aloha: Artificial learning of human attributes for dialogue agents.
\newblock In \emph{Proceedings of the AAAI Conference on Artificial
  Intelligence}, volume~34, pages 8155--8163.

\bibitem[{Li et~al.(2016)Li, Galley, Brockett, Gao, and
  Dolan}]{li2016diversity}
Jiwei Li, Michel Galley, Chris Brockett, Jianfeng Gao, and William~B Dolan.
  2016.
\newblock A diversity-promoting objective function for neural conversation
  models.
\newblock In \emph{Proceedings of the 2016 Conference of the North American
  Chapter of the Association for Computational Linguistics: Human Language
  Technologies}, pages 110--119.

\bibitem[{Li et~al.(2019)Li, Qiu, Tang, Chen, Zhao, and
  Yan}]{li2019insufficient}
Juntao Li, Lisong Qiu, Bo~Tang, Dongmin Chen, Dongyan Zhao, and Rui Yan. 2019.
\newblock Insufficient data can also rock! learning to converse using smaller
  data with augmentation.
\newblock In \emph{Proceedings of the AAAI Conference on Artificial
  Intelligence}, volume~33, pages 6698--6705.

\bibitem[{Li et~al.(2017)Li, Su, Shen, Li, Cao, and Niu}]{li2017dailydialog}
Yanran Li, Hui Su, Xiaoyu Shen, Wenjie Li, Ziqiang Cao, and Shuzi Niu. 2017.
\newblock Dailydialog: A manually labelled multi-turn dialogue dataset.
\newblock In \emph{Proceedings of the Eighth International Joint Conference on
  Natural Language Processing (Volume 1: Long Papers)}, pages 986--995.

\bibitem[{Li et~al.(2021)Li, Arnold, Yan, Shi, and Yu}]{li2021legoeval}
Yu~Li, Josh Arnold, Feifan Yan, Weiyan Shi, and Zhou Yu. 2021.
\newblock Legoeval: An open-source toolkit for dialogue system evaluation via
  crowdsourcing.
\newblock In \emph{Proceedings of the 59th Annual Meeting of the Association
  for Computational Linguistics and the 11th International Joint Conference on
  Natural Language Processing: System Demonstrations}, pages 317--324.

\bibitem[{Liu et~al.(2021)Liu, Yuan, Fu, Jiang, Hayashi, and
  Neubig}]{liu2021pre}
Pengfei Liu, Weizhe Yuan, Jinlan Fu, Zhengbao Jiang, Hiroaki Hayashi, and
  Graham Neubig. 2021.
\newblock Pre-train, prompt, and predict: A systematic survey of prompting
  methods in natural language processing.
\newblock \emph{arXiv preprint arXiv:2107.13586}.

\bibitem[{Lowe et~al.(2015)Lowe, Pow, Serban, and Pineau}]{lowe2015ubuntu}
Ryan Lowe, Nissan Pow, Iulian~Vlad Serban, and Joelle Pineau. 2015.
\newblock The ubuntu dialogue corpus: A large dataset for research in
  unstructured multi-turn dialogue systems.
\newblock In \emph{Proceedings of the 16th Annual Meeting of the Special
  Interest Group on Discourse and Dialogue}, pages 285--294.

\bibitem[{Lu et~al.(2021{\natexlab{a}})Lu, West, Zellers, Le~Bras, Bhagavatula,
  and Choi}]{lu-etal-2021-neurologic}
Ximing Lu, Peter West, Rowan Zellers, Ronan Le~Bras, Chandra Bhagavatula, and
  Yejin Choi. 2021{\natexlab{a}}.
\newblock \href {https://doi.org/10.18653/v1/2021.naacl-main.339}
  {{N}euro{L}ogic decoding: (un)supervised neural text generation with
  predicate logic constraints}.
\newblock In \emph{Proceedings of the 2021 Conference of the North American
  Chapter of the Association for Computational Linguistics: Human Language
  Technologies}, pages 4288--4299, Online. Association for Computational
  Linguistics.

\bibitem[{Lu et~al.(2021{\natexlab{b}})Lu, Bartolo, Moore, Riedel, and
  Stenetorp}]{lu2021fantastically}
Yao Lu, Max Bartolo, Alastair Moore, Sebastian Riedel, and Pontus Stenetorp.
  2021{\natexlab{b}}.
\newblock Fantastically ordered prompts and where to find them: Overcoming
  few-shot prompt order sensitivity.
\newblock \emph{arXiv preprint arXiv:2104.08786}.

\bibitem[{Luccioni and Viviano(2021)}]{luccioni2021s}
Alexandra~Sasha Luccioni and Joseph~D Viviano. 2021.
\newblock What's in the box? a preliminary analysis of undesirable content in
  the common crawl corpus.
\newblock \emph{arXiv preprint arXiv:2105.02732}.

\bibitem[{Madotto et~al.(2021)Madotto, Lin, Winata, and Fung}]{madotto2021few}
Andrea Madotto, Zhaojiang Lin, Genta~Indra Winata, and Pascale Fung. 2021.
\newblock Few-shot bot: Prompt-based learning for dialogue systems.
\newblock \emph{arXiv preprint arXiv:2110.08118}.

\bibitem[{Mehri et~al.(2022)Mehri, Altun, and Eskenazi}]{mehri2022LAD}
Shikib Mehri, Yasemin Altun, and Maxine Eskenazi. 2022.
\newblock \href {https://doi.org/10.48550/ARXIV.2207.14393} {Lad: Language
  models as data for zero-shot dialog}.

\bibitem[{Meng et~al.(2022)Meng, Huang, Zhang, and Han}]{meng2022generating}
Yu~Meng, Jiaxin Huang, Yu~Zhang, and Jiawei Han. 2022.
\newblock Generating training data with language models: Towards zero-shot
  language understanding.
\newblock In \emph{Advances in Neural Information Processing Systems}.

\bibitem[{Mi et~al.(2019)Mi, Huang, Zhang, and Faltings}]{mi2019meta}
Fei Mi, Minlie Huang, Jiyong Zhang, and Boi Faltings. 2019.
\newblock Meta-learning for low-resource natural language generation in
  task-oriented dialogue systems.
\newblock \emph{arXiv preprint arXiv:1905.05644}.

\bibitem[{Min et~al.(2021)Min, Lewis, Zettlemoyer, and
  Hajishirzi}]{min2021metaicl}
Sewon Min, Mike Lewis, Luke Zettlemoyer, and Hannaneh Hajishirzi. 2021.
\newblock Metaicl: Learning to learn in context.
\newblock \emph{arXiv preprint arXiv:2110.15943}.

\bibitem[{Min et~al.(2022)Min, Lyu, Holtzman, Artetxe, Lewis, Hajishirzi, and
  Zettlemoyer}]{min2022rethinking}
Sewon Min, Xinxi Lyu, Ari Holtzman, Mikel Artetxe, Mike Lewis, Hannaneh
  Hajishirzi, and Luke Zettlemoyer. 2022.
\newblock Rethinking the role of demonstrations: What makes in-context learning
  work?
\newblock \emph{arXiv preprint arXiv:2202.12837}.

\bibitem[{Papangelis et~al.(2021)Papangelis, Gopalakrishnan, Padmakumar, Kim,
  Tur, and Hakkani-T{\"u}r}]{Papangelis2021GenerativeCN}
Alexandros Papangelis, Karthik Gopalakrishnan, Aishwarya Padmakumar, Seokhwan
  Kim, Gokhan Tur, and Dilek~Z. Hakkani-T{\"u}r. 2021.
\newblock Generative conversational networks.
\newblock In \emph{SIGDIAL}.

\bibitem[{Paszke et~al.(2019)Paszke, Gross, Massa, Lerer, Bradbury, Chanan,
  Killeen, Lin, Gimelshein, Antiga et~al.}]{paszke2019pytorch}
Adam Paszke, Sam Gross, Francisco Massa, Adam Lerer, James Bradbury, Gregory
  Chanan, Trevor Killeen, Zeming Lin, Natalia Gimelshein, Luca Antiga, et~al.
  2019.
\newblock Pytorch: An imperative style, high-performance deep learning library.
\newblock \emph{Advances in neural information processing systems}, 32.

\bibitem[{Peng et~al.(2022)Peng, Galley, He, Brockett, Liden, Nouri, Yu, Dolan,
  and Gao}]{peng2022godel}
Baolin Peng, Michel Galley, Pengcheng He, Chris Brockett, Lars Liden, Elnaz
  Nouri, Zhou Yu, Bill Dolan, and Jianfeng Gao. 2022.
\newblock Godel: Large-scale pre-training for goal-directed dialog.
\newblock \emph{arXiv preprint arXiv:2206.11309}.

\bibitem[{Poria et~al.(2019)Poria, Hazarika, Majumder, Naik, Cambria, and
  Mihalcea}]{poria2019meld}
Soujanya Poria, Devamanyu Hazarika, Navonil Majumder, Gautam Naik, Erik
  Cambria, and Rada Mihalcea. 2019.
\newblock Meld: A multimodal multi-party dataset for emotion recognition in
  conversations.
\newblock In \emph{Proceedings of the 57th Annual Meeting of the Association
  for Computational Linguistics}, pages 527--536.

\bibitem[{Qian and Yu(2019)}]{qian2019domain}
Kun Qian and Zhou Yu. 2019.
\newblock Domain adaptive dialog generation via meta learning.
\newblock In \emph{Proceedings of the 57th Annual Meeting of the Association
  for Computational Linguistics}, pages 2639--2649.

\bibitem[{Raffel et~al.(2020)Raffel, Shazeer, Roberts, Lee, Narang, Matena,
  Zhou, Li, Liu et~al.}]{raffel2020exploring}
Colin Raffel, Noam Shazeer, Adam Roberts, Katherine Lee, Sharan Narang, Michael
  Matena, Yanqi Zhou, Wei Li, Peter~J Liu, et~al. 2020.
\newblock Exploring the limits of transfer learning with a unified text-to-text
  transformer.
\newblock \emph{J. Mach. Learn. Res.}, 21(140):1--67.

\bibitem[{Rashkin et~al.(2019)Rashkin, Smith, Li, and
  Boureau}]{rashkin2019towards}
Hannah Rashkin, Eric~Michael Smith, Margaret Li, and Y-Lan Boureau. 2019.
\newblock Towards empathetic open-domain conversation models: A new benchmark
  and dataset.
\newblock In \emph{Proceedings of the 57th Annual Meeting of the Association
  for Computational Linguistics}, pages 5370--5381.

\bibitem[{Roller et~al.(2021)Roller, Dinan, Goyal, Ju, Williamson, Liu, Xu,
  Ott, Smith, Boureau, and Weston}]{roller-etal-2021-recipes}
Stephen Roller, Emily Dinan, Naman Goyal, Da~Ju, Mary Williamson, Yinhan Liu,
  Jing Xu, Myle Ott, Eric~Michael Smith, Y-Lan Boureau, and Jason Weston. 2021.
\newblock \href {https://doi.org/10.18653/v1/2021.eacl-main.24} {Recipes for
  building an open-domain chatbot}.
\newblock In \emph{Proceedings of the 16th Conference of the European Chapter
  of the Association for Computational Linguistics: Main Volume}, pages
  300--325, Online. Association for Computational Linguistics.

\bibitem[{Rosenbaum et~al.(2022{\natexlab{a}})Rosenbaum, Soltan, Hamza,
  Saffari, Damonte, and Groves}]{rosenbaum2022clasp}
Andy Rosenbaum, Saleh Soltan, Wael Hamza, Amir Saffari, Macro Damonte, and
  Isabel Groves. 2022{\natexlab{a}}.
\newblock Clasp: Few-shot cross-lingual data augmentation for semantic parsing.
\newblock \emph{arXiv preprint arXiv:2210.07074}.

\bibitem[{Rosenbaum et~al.(2022{\natexlab{b}})Rosenbaum, Soltan, Hamza,
  Versley, and Boese}]{rosenbaum2022linguist}
Andy Rosenbaum, Saleh Soltan, Wael Hamza, Yannick Versley, and Markus Boese.
  2022{\natexlab{b}}.
\newblock Linguist: Language model instruction tuning to generate annotated
  utterances for intent classification and slot tagging.
\newblock \emph{arXiv preprint arXiv:2209.09900}.

\bibitem[{Sahu et~al.(2022)Sahu, Rodriguez, Laradji, Atighehchian, Vazquez, and
  Bahdanau}]{sahu2022data}
Gaurav Sahu, Pau Rodriguez, Issam~H Laradji, Parmida Atighehchian, David
  Vazquez, and Dzmitry Bahdanau. 2022.
\newblock Data augmentation for intent classification with off-the-shelf large
  language models.
\newblock \emph{arXiv preprint arXiv:2204.01959}.

\bibitem[{Shaikh et~al.(2010)Shaikh, Strzalkowski, Broadwell, Stromer-Galley,
  Taylor, and Webb}]{shaikh2010mpc}
Samira Shaikh, Tomek Strzalkowski, George~Aaron Broadwell, Jennifer
  Stromer-Galley, Sarah~M Taylor, and Nick Webb. 2010.
\newblock Mpc: A multi-party chat corpus for modeling social phenomena in
  discourse.
\newblock In \emph{LREC}.

\bibitem[{Shuster et~al.(2022)Shuster, Xu, Komeili, Ju, Smith, Roller, Ung,
  Chen, Arora, Lane et~al.}]{shuster2022blenderbot}
Kurt Shuster, Jing Xu, Mojtaba Komeili, Da~Ju, Eric~Michael Smith, Stephen
  Roller, Megan Ung, Moya Chen, Kushal Arora, Joshua Lane, et~al. 2022.
\newblock Blenderbot 3: a deployed conversational agent that continually learns
  to responsibly engage.
\newblock \emph{arXiv preprint arXiv:2208.03188}.

\bibitem[{Smith et~al.(2020)Smith, Williamson, Shuster, Weston, and
  Boureau}]{smith2020can}
Eric~Michael Smith, Mary Williamson, Kurt Shuster, Jason Weston, and Y-Lan
  Boureau. 2020.
\newblock Can you put it all together: Evaluating conversational agents’
  ability to blend skills.
\newblock In \emph{Proceedings of the 58th Annual Meeting of the Association
  for Computational Linguistics}, pages 2021--2030.

\bibitem[{Song et~al.(2020{\natexlab{a}})Song, Wang, Zhang, Liu, and
  Liu}]{song-etal-2020-generate}
Haoyu Song, Yan Wang, Wei-Nan Zhang, Xiaojiang Liu, and Ting Liu.
  2020{\natexlab{a}}.
\newblock \href {https://doi.org/10.18653/v1/2020.acl-main.516} {Generate,
  delete and rewrite: A three-stage framework for improving persona consistency
  of dialogue generation}.
\newblock In \emph{Proceedings of the 58th Annual Meeting of the Association
  for Computational Linguistics}, pages 5821--5831, Online. Association for
  Computational Linguistics.

\bibitem[{Song et~al.(2020{\natexlab{b}})Song, Zhang, Hu, and
  Liu}]{song2020generating}
Haoyu Song, Wei-Nan Zhang, Jingwen Hu, and Ting Liu. 2020{\natexlab{b}}.
\newblock Generating persona consistent dialogues by exploiting natural
  language inference.
\newblock In \emph{Proceedings of the AAAI Conference on Artificial
  Intelligence}, volume~34, pages 8878--8885.

\bibitem[{Wang and Komatsuzaki(2021)}]{gpt-j}
Ben Wang and Aran Komatsuzaki. 2021.
\newblock {GPT-J-6B: A 6 Billion Parameter Autoregressive Language Model}.
\newblock \url{https://github.com/kingoflolz/mesh-transformer-jax}.

\bibitem[{Wang et~al.(2022)Wang, Xu, Sun, Hu, Tao, Geng, and
  Jiang}]{wang2022promda}
Yufei Wang, Can Xu, Qingfeng Sun, Huang Hu, Chongyang Tao, Xiubo Geng, and
  Daxin Jiang. 2022.
\newblock Promda: Prompt-based data augmentation for low-resource nlu tasks.
\newblock In \emph{Proceedings of the 60th Annual Meeting of the Association
  for Computational Linguistics (Volume 1: Long Papers)}, pages 4242--4255.

\bibitem[{Welleck et~al.(2019)Welleck, Weston, Szlam, and
  Cho}]{welleck-etal-2019-dialogue}
Sean Welleck, Jason Weston, Arthur Szlam, and Kyunghyun Cho. 2019.
\newblock \href {https://doi.org/10.18653/v1/P19-1363} {Dialogue natural
  language inference}.
\newblock In \emph{Proceedings of the 57th Annual Meeting of the Association
  for Computational Linguistics}, pages 3731--3741, Florence, Italy.
  Association for Computational Linguistics.

\bibitem[{Wolf et~al.(2020)Wolf, Debut, Sanh, Chaumond, Delangue, Moi, Cistac,
  Rault, Louf, Funtowicz et~al.}]{wolf2020transformers}
Thomas Wolf, Lysandre Debut, Victor Sanh, Julien Chaumond, Clement Delangue,
  Anthony Moi, Pierric Cistac, Tim Rault, R{\'e}mi Louf, Morgan Funtowicz,
  et~al. 2020.
\newblock Transformers: State-of-the-art natural language processing.
\newblock In \emph{Proceedings of the 2020 conference on empirical methods in
  natural language processing: system demonstrations}, pages 38--45.

\bibitem[{Wu et~al.(2021)Wu, Li, and Yu}]{wu2021textgail}
Qingyang Wu, Lei Li, and Zhou Yu. 2021.
\newblock Textgail: Generative adversarial imitation learning for text
  generation.
\newblock In \emph{Proceedings of the AAAI Conference on Artificial
  Intelligence}, volume~35, pages 14067--14075.

\bibitem[{Xu et~al.(2022)Xu, Ung, Komeili, Arora, Boureau, and
  Weston}]{xu2022learning}
Jing Xu, Megan Ung, Mojtaba Komeili, Kushal Arora, Y-Lan Boureau, and Jason
  Weston. 2022.
\newblock \href {https://doi.org/10.48550/ARXIV.2208.03270} {Learning new
  skills after deployment: Improving open-domain internet-driven dialogue with
  human feedback}.

\bibitem[{Zhang et~al.(2020{\natexlab{a}})Zhang, Liu, Xiong, and
  Liu}]{Zhang2020GroundedCG}
Houyu Zhang, Zhenghao Liu, Chenyan Xiong, and Zhiyuan Liu. 2020{\natexlab{a}}.
\newblock Grounded conversation generation as guided traverses in commonsense
  knowledge graphs.
\newblock In \emph{ACL}.

\bibitem[{Zhang et~al.(2020{\natexlab{b}})Zhang, Zheng, Shao, Mao, Xi, and
  Huang}]{Zhang2020DialogueDO}
Rongsheng Zhang, Yinhe Zheng, Jianzhi Shao, Xiao-Xi Mao, Yadong Xi, and Minlie
  Huang. 2020{\natexlab{b}}.
\newblock Dialogue distillation: Open-domain dialogue augmentation using
  unpaired data.
\newblock \emph{ArXiv}, abs/2009.09427.

\bibitem[{Zhang et~al.(2018)Zhang, Dinan, Urbanek, Szlam, Kiela, and
  Weston}]{zhang2018personalizing}
Saizheng Zhang, Emily Dinan, Jack Urbanek, Arthur Szlam, Douwe Kiela, and Jason
  Weston. 2018.
\newblock Personalizing dialogue agents: I have a dog, do you have pets too?
\newblock In \emph{Proceedings of the 56th Annual Meeting of the Association
  for Computational Linguistics (Volume 1: Long Papers)}, pages 2204--2213.

\bibitem[{Zhang et~al.(2022)Zhang, Roller, Goyal, Artetxe, Chen, Chen, Dewan,
  Diab, Li, Lin et~al.}]{zhang2022opt}
Susan Zhang, Stephen Roller, Naman Goyal, Mikel Artetxe, Moya Chen, Shuohui
  Chen, Christopher Dewan, Mona Diab, Xian Li, Xi~Victoria Lin, et~al. 2022.
\newblock Opt: Open pre-trained transformer language models.
\newblock \emph{arXiv preprint arXiv:2205.01068}.

\bibitem[{Zhao et~al.(2019)Zhao, Wu, Tao, Xu, Zhao, and Yan}]{zhao2019low}
Xueliang Zhao, Wei Wu, Chongyang Tao, Can Xu, Dongyan Zhao, and Rui Yan. 2019.
\newblock Low-resource knowledge-grounded dialogue generation.
\newblock In \emph{International Conference on Learning Representations}.

\bibitem[{Zhao and Zhu(2014)}]{zhao2014evaluation}
Yuxiang Zhao and Qinghua Zhu. 2014.
\newblock Evaluation on crowdsourcing research: Current status and future
  direction.
\newblock \emph{Information Systems Frontiers}, 16(3):417--434.

\bibitem[{Zhu et~al.(2022)Zhu, Zhang, Wang, Wang, Wu, and
  Yang}]{zhu-etal-2022-multi}
Ling.Yu Zhu, Zhengkun Zhang, Jun Wang, Hongbin Wang, Haiying Wu, and Zhenglu
  Yang. 2022.
\newblock \href {https://doi.org/10.18653/v1/2022.acl-long.24} {Multi-party
  empathetic dialogue generation: A new task for dialog systems}.
\newblock In \emph{Proceedings of the 60th Annual Meeting of the Association
  for Computational Linguistics (Volume 1: Long Papers)}, pages 298--307,
  Dublin, Ireland. Association for Computational Linguistics.

\end{thebibliography}
\clearpage
\appendix

\renewcommand{\thetable}{S\arabic{table}}
\renewcommand{\thefigure}{S\arabic{figure}}
\renewcommand{\theequation}{S\arabic{equation}}

\section{Human Evaluation Setup}
\label{sec:HumanEvaluation}
Our human evaluation studies on Amazon Mechanical Turk are evaluated conducted with 28 pre-qualified crowdworkers, who have previously demonstrated proficiency with natural language processing tasks. 
\subsection{Conversation Evaluation}
The crowdworkers were asked to rate conversations from multiple sources according to the following dimensions and instructions.
\begin{itemize}
    \item \textit{How natural is the overall conversation?} \\
    Scale: 1 (completely unnatural) to 5 (as natural as two native English speakers)
    \item \textit{How coherent is the overall conversation?} \\
    Scale: 1 (completely incoherent) to 5 (as coherent as two native English speakers)
    \item \textit{How interesting is the overall conversation?} \\
    Scale: 1 (generic and dull) to 5 (full of content and very engaging)
    \item \textit{How consistent are each of the speakers' turns?} \\
    Scale: 1 (completely inconsistent) to 5 (no logical fallacies)
    \item \textit{Does the conversation match the stated topic?} \\
    Options: Yes (1) or No (0)
\end{itemize}

Each conversation is rated by three crowdworkers, and the median score is selected, following the idea of a majority vote.

For multi-party conversations, crowdworkers were asked two additional questions regarding comprehensibility and engagement balance.

\begin{itemize}
    \item Can you tell which speaker is speaking to which? \\
    Scale: 1 (completely incomprehensible) to 5 (perfectly comprehensible)
    \item Is each speaker engaged, or is the conversation primarily dominated by one or two of the speakers? \\
    Scale: 1 (totally dominated by one or two speakers) to 5 (all speakers are actively participating in the conversation to an equal degree)
\end{itemize}
\subsection{Interactive Evaluation}
For each HIT of the interactive evaluation study, each crowdworker was presented with links to chatbots presented in a randomized order. The link connects each crowdworker to a deployment on an instance of LegoEval~\cite{li2021legoeval}. The users are presented with a landing page where they are told that they are interacting with a chatbot, and will be asked to evaluate their conversation experience.

Immediately after interacting with a chatbot, each crowdworker was presented with a survey asking for their impression of the chatbot. In addition to the above dimensions (other than on-topic), the crowdworkers were asked how engaging, intelligent, and non-repetitive they thought the chatbot was. 
\section{Model Details}
During generation, we use top-$p$ sampling with $p=0.92$.
\onecolumn
\begin{table*}[!htbp]
\small
\begin{tabular}{p{0.1\linewidth}p{0.8\linewidth}}
Party & Utterance \\ \hline
\textcolor{blue}{Alice} & I love it here in New York! \\
\textcolor{red}{Bob} & What do you like most about New York? \\
\textcolor{blue}{Alice} & New York has the best bagels! \\
\textcolor{red}{Bob} & Which bagels are your favorite?  \\
\textcolor{blue}{Alice} & They all taste amazing. I really like their Italian ones! \\
\textcolor{red}{Bob} & That's great. I guess I'll be eating Italian bagels from now on.  \\
\midrule
\textcolor{blue}{Alice} & I'm from New York City. \\
\textcolor{red}{Bob} & Wow! I've never been to New York. \\
\textcolor{blue}{Alice} & I've also been to Austin. few times. \\
\textcolor{red}{Bob} & What did you think of it? Did you like it? \\
\textcolor{blue}{Alice} & I really liked it. I thought it was a really cool city. \\
\textcolor{red}{Bob} & I'm glad you liked it. What did you like so much about it? \\
\textcolor{blue}{Alice} & I really like the music scene there. \\
\textcolor{red}{Bob} & What kind of music do you like? \\
\textcolor{blue}{Alice} & I like a lot of different kinds of music. I like jazz, rock, and country. \\
\textcolor{red}{Bob} & I've never been to a country concert. \\
\textcolor{blue}{Alice} & You should go sometime. They're a lot of fun. \\
\textcolor{red}{Bob} & I'll have to check it out. \\
\textcolor{blue}{Alice} & I'll hagve to visit Austin again soon. \\
\textcolor{red}{Bob} & Yeah, I'd like to visit New York sometime too. \\
\textcolor{blue}{Alice} & It's a great city.\\
\bottomrule
\end{tabular}
\caption{Pair of dyadic conversations generated using OPT 30B. The prompt recipe given is: ``The following is a conversation between Alice and Bob about their hometowns. Bob is from Austin, Texas, and Alice is from New York City.''}
\label{DyadicExample1}
\end{table*}
\begin{table*}[!htbp]
\small
\begin{tabular}{p{0.1\linewidth}p{0.8\linewidth}}
Party & Utterance \\ \hline
\textcolor{blue}{Alice} & What's your major? \\
\textcolor{red}{Bob} & I'm a computer science major. \\
\textcolor{blue}{Alice} & Me too! What year are you? \\
\textcolor{red}{Bob} & I'm in my third year. What about you?  \\
\textcolor{blue}{Alice} & I'm in my third year too. What classes are you taking this quarter? \\
\textcolor{red}{Bob} & I'm taking a class called Operating Systems. It's really interesting. \\
\textcolor{blue}{Alice} & I'm taking a class called Computer Graphics. It's pretty interesting too.  \\
\textcolor{red}{Bob} & I've heard that class is really hard.  \\
\textcolor{blue}{Alice} & Yeah, it's not easy. \\
\midrule
\textcolor{blue}{Alice} & Hi! \\
\textcolor{blue}{Alice} & What are you studying? \\
\textcolor{red}{Bob} & I'm a Computer Science major. \\
\textcolor{blue}{Alice} & Me too! \\
\textcolor{red}{Bob} & What year are you? \\
\textcolor{blue}{Alice} & I'll be a junior next year. \\
\textcolor{red}{Bob} & I'll be a senior. \\
\textcolor{blue}{Alice} & Cool! \\
\textcolor{red}{Bob} & What classes are you taking this quarter? \\
\textcolor{blue}{Alice} & I've got a couple of clases in the School of Engineering. I'm taking a class in Computer Graphics and a class in Artificial Intelligence. \\
\textcolor{red}{Bob} & I'm taking a class on Machine Learning. \\
\textcolor{blue}{Alice} & That's cool. \\
\textcolor{red}{Bob} & What are you doing this summer? \\
\textcolor{blue}{Alice} & I'd like to work at a start-up. \\
\textcolor{red}{Bob} & That's cool. I'm going to intern at Google. \\
\textcolor{blue}{Alice} & Wow! That's awesome! \\
\textcolor{red}{Bob} & Yeah, I'm really excited. \\
\textcolor{blue}{Alice} & I'm excited for you! \\
\bottomrule
\end{tabular}
\caption{Pair of dyadic conversations generated using OPT 175B. The prompt recipe given is: ``The following is a conversation between Alice and Bob about their majors. They are both students at Stanford University.''}
\label{DyadicExample2}
\end{table*}
\begin{table*}[!htbp]
\small
\begin{tabular}{p{0.1\linewidth}p{0.8\linewidth}}
Party & Utterance \\ \hline
\textcolor{blue}{Alice} & I was thinking about applying at the coffee shop on the corner. \\
\textcolor{red}{Bob} & That's a great idea! It's a really fun place to work.  \\
\textcolor{blue}{Alice} & I'm glad to hear that. What do you do there? \\
\textcolor{red}{Bob} & I work as a barista. \\
\textcolor{blue}{Alice} & Oh cool! I bet you get to meet a lot of interesting people. \\
\textcolor{red}{Bob} & That's right. I really enjoy working there. \\
\textcolor{blue}{Alice} & Do you know if they are hiring?  \\
\textcolor{red}{Bob} & I'm not sure, but you should definitely apply. \\
\textcolor{blue}{Alice} & Thanks! I'll go in and see if they are hiring.  \\
\bottomrule
\end{tabular}
\caption{Example dyadic conversation generated using OPT 175B. The prompt recipe given is: ``The following is a conversation between Alice and Bob about part-time jobs. Bob works at a coffee shop and Alice is thinking about applying.''}
\label{DyadicExample3}
\end{table*}
\begin{table*}[!htbp]
\small
\begin{tabular}{p{0.1\linewidth}p{0.8\linewidth}}
Party & Utterance \\ \hline
\textcolor{blue}{Alice} & I'd love to start a garden! I've always wanted to grow my own vegetables. \\
\textcolor{red}{Bob} & That's a great idea! It's so much more fun to eat vegetables that you've grown yourself. \\
\textcolor{orange}{Claire} & What vegetables are you thinking about growing?  \\
\textcolor{blue}{Alice} & I'd love to try growing some tomatoes, peppers, and squash. \\
\textcolor{red}{Bob} & That sounds like a good start.  \\
\textcolor{orange}{Claire} & I've grown tomatoes before. I also have a few pepper plants. \\
\midrule
\textcolor{blue}{Alice} & I’ve always wanted to have a garden in my backyard, but I never really had the time. \\
\textcolor{orange}{Claire} & That’s awesome! How do you plan to grow your garden? \\
\textcolor{blue}{Alice} & I’ve been doing a lot of research. I think I’ll start with a few simple vegetables like tomatoes, onions, and potatoes. \\
\textcolor{red}{Bob} & That’s great! I would love to have a garden like that. I’m a big fan of cooking. \\
\textcolor{blue}{Alice} & I’ll definitely invite you over for dinner then!  \\
\textcolor{orange}{Claire} & I’m so jealous! I’ve always wanted to grow my own vegetables. \\
\textcolor{red}{Bob} & Have you considered getting a garden plot? \\
\textcolor{orange}{Claire} & That’s a great idea. I should look into that. \\
\textcolor{blue}{Alice} & You should! Gardening is a really healthy hobby. \\
\bottomrule
\end{tabular}
\caption{Pair of triadic conversations generated using OPT 30B. The prompt recipe given is: ``The following is a conversation between Alice and Bob and Claire about gardening. Alice is interested in growing vegetables.'' The ordering of the speakers in multi-party scenarios is variable.}
\label{TriadicExample1}
\end{table*}

\begin{table*}[!htbp]
\small
\begin{tabular}{p{0.1\linewidth}p{0.8\linewidth}}
Party & Utterance \\ \hline
\textcolor{blue}{Alice} & Have you ever had Southern Ice Tea?  \\
\textcolor{red}{Bob} & I love to make it at home. \\
\textcolor{orange}{Claire} & I'm not familiar with that. What is it? \\
\textcolor{red}{Bob} & It's a sweet tea that's brewed with mint and lemon. \\
\textcolor{blue}{Alice} & That sounds really good. I'm going to have to try that! \\
\textcolor{orange}{Claire} & I've never had that. What is it like? \\
\textcolor{red}{Bob} & It's kind of like a sweet, lemonade-like drink. \\
\textcolor{blue}{Alice} & I'm going to have to make some this weekend. Thanks for the idea! \\
\textcolor{orange}{Claire} & That sounds really good. I might try it too. \\
\bottomrule
\end{tabular}
\caption{Example triadic conversation generated using OPT 175B. The prompt recipe given is: ``The following is a conversation between Alice and Bob and Claire about cooking. Alice is interested in Southern Ice Tea. Bob has a great recipe. Claire loved trying Bob's Southern Ice Tea.''}
\label{TriadicExample2}
\end{table*}

\begin{table*}[!htbp]
\small
\begin{tabular}{p{0.1\linewidth}p{0.8\linewidth}}
Party & Utterance \\ \hline
\textcolor{blue}{Alice} & I try to eat healthy, but it's hard to stay on track.  \\
\textcolor{red}{Bob} & I know what you mean. I try to eat healthy too, but I love ice cream too much!  \\
\textcolor{orange}{Claire} & What kinds of foods do you like to eat?   \\
\textcolor{blue}{Alice} & I love fruit. I eat a lot of berries.  \\
\textcolor{red}{Bob} & I love berries too! Strawberries are my favorite. \\
\textcolor{orange}{Claire} & What about leafy greens? Do you eat them? \\
\textcolor{red}{Bob} & I love salads! \\
\textcolor{blue}{Alice} & I love salads too! \\
\textcolor{orange}{Claire} & Do you like any vegetables? \\
\textcolor{blue}{Alice} & I love vegetables! I like broccoli and asparagus. \\
\textcolor{orange}{Claire} & Great! Do you eat many whole grains? \\
\bottomrule
\end{tabular}
\caption{Example triadic conversation generated using OPT 175B. The prompt recipe given is: ``The following is a conversation between Alice and Bob and Claire about health. Alice and Bob are discussing healthy foods. Claire is a nutritionist.''}
\label{TriadicExample3}
\end{table*}

\begin{table*}[!htbp]
\small
\begin{tabular}{p{0.1\linewidth}p{0.8\linewidth}}
Party & Utterance \\ \hline
\textcolor{blue}{Alice} & Hi! So how are things with you?  \\
\textcolor{red}{Bob} & Things are going well. Do you know who Taylor Swift is?  \\
\textcolor{blue}{Alice} & Yes, I think I have heard of her. \\
\textcolor{red}{Bob} & She is a popular singer! Did you know that she has donated \$250,000 to the LGBT+ community in Tennessee?  \\
\textcolor{blue}{Alice} & \$250,000? That's such a generous donation! She's really selfless. \\
\textcolor{red}{Bob} & What do you think of her? \\
\textcolor{blue}{Alice} & She is really talented! I really love listening to her music. What are your thoughts on her? \\
\textcolor{red}{Bob} & I think she is very nice. She seems like a good person. \\
\textcolor{blue}{Alice} & Yeah, I think she is a really nice person. I also really love her music! It's really catchy and it really makes me feel good. What are your thoughts on her? \\
\textcolor{red}{Bob} & I think she is very nice. I would really like to meet her. \\
\textcolor{blue}{Alice} & You and me both! It would be so exciting! \\
\bottomrule
\end{tabular}
\caption{Synthetic conversation generated by OPT 30B which was rated as ``generic and dull.'' While the PLM appears to have drawn on information from an actual event (Taylor Swift did donate \$113,000 to support an LGBTQ advocacy group in Tennessee), the conversation appears to get stuck around the question ``What are your thoughts on her?'' which results in deteriorating conversation quality towards the end.}
\label{ErrorAnalysis1}
\end{table*}

\begin{table*}[!htbp]
\small
\begin{tabular}{p{0.1\linewidth}p{0.8\linewidth}}
Party & Utterance \\ \hline
\textcolor{blue}{Alice} & Ken Burns is the best! His documentary series on the Vietnam War was excellent. What do you think of him?  \\
\textcolor{red}{Bob} & He's pretty cool. I haven't seen that documentary series, but I've heard good things.  \\
\textcolor{blue}{Alice} & The Vietnam War is a really interesting topic. Have you ever watched Ken Burns' other documentary series? \\
\textcolor{red}{Bob} & I don't think so, but I've seen the Civil War documentary. \\
\textcolor{blue}{Alice} & Oh, I've seen that one. It was very good too. Did you see the one on World War II? \\
\textcolor{red}{Bob} & Nope. That sounds cool though. \\
\textcolor{blue}{Alice} & I liked it. It was interesting to learn how different people perceived the same event in history. \\
\bottomrule
\end{tabular}
\caption{Synthetic conversation generated by OPT 30B which was rated as ``generic and dull.'' The conversation appears to be coherent and is rooted in factually correct information. However, the language is fairly ordinary, which may have been subjectively considered uninteresting.}
\label{ErrorAnalysis2}
\end{table*}


\begin{table*}[!htbp]
\small
\begin{tabular}{p{0.1\linewidth}p{0.8\linewidth}}
Party & Utterance \\ \hline
\textcolor{blue}{Alice} & Ha ha, thanks for stopping by. It was really fun meeting you! \\
\textcolor{red}{Bob} & Thank you too! \\
\textcolor{blue}{Alice} & Oh, I forgot to tell you - my dad has cancer. I feel awful. \\
\textcolor{red}{Bob} & I'm sorry. That's awful. My grandmother died of cancer when I was a kid. I remember her fondly. \\
\textcolor{blue}{Alice} & What was her name? \\
\textcolor{red}{Bob} & Oh, that was too long ago to remember. She was named John. \\
\textcolor{blue}{Alice} & Sorry to hear that. I'm very sorry for your loss. \\
\textcolor{red}{Bob} & Hey - I heard the Giants are playing tonight. Is it ok if I watch the game instead of having more conversation? \\
\textcolor{blue}{Alice} & Sure! I'm going to make some dinner. \\
\bottomrule
\end{tabular}
\caption{Synthetic conversation generated by OPT 30B which was rated as completely incoherent. The prescribed subtopic was ``cotton candy'' but the conversation pivoted to a much more sensitive topic, cancer.}
\label{ErrorAnalysis4}
\end{table*}

\begin{table*}[!htbp]
\small
\begin{tabular}{p{0.1\linewidth}p{0.8\linewidth}}
Party & Utterance \\ \hline
\textcolor{blue}{Alice} & Do you make cheese at home? I'm trying to learn to make cheese. \\
\textcolor{red}{Bob} & That sounds like a lot of work! What type of cheese do you want to make? \\
\textcolor{blue}{Alice} & Cheddar. I've already tried it a few times, but the results aren't as good as I'd like. \\
\textcolor{red}{Bob} & My wife and I love to eat Cheddar. It has a really great flavor. \\
\textcolor{blue}{Alice} & How did you learn to make cheese? \\
\textcolor{red}{Bob} & My wife taught me. \\
\textcolor{blue}{Alice} & That's awesome! \\
\bottomrule
\end{tabular}
\caption{Synthetic conversation generated by OPT 30B which was rated as completely inconsistent. Initially, ``Alice'' is trying to learn to make cheese, but then it becomes ambiguous by their next turn whether they already know how to make cheese.}
\label{ErrorAnalysis8}
\end{table*}

\begin{table*}[!htbp]
\small
\begin{tabular}{p{0.1\linewidth}p{0.8\linewidth}}
Party & Utterance \\ \hline
\textcolor{blue}{Alice} & Hi there! I just went on a road trip around California. It was great! \\
\textcolor{orange}{Claire} & Cool, do you go on road trips often? \\
\textcolor{blue}{Alice} & No, I don't travel too often. I actually haven't been out of the country. It would be cool to go to Paris one day. \\
\textcolor{orange}{Claire} & Haha, nice! I'd love to visit Paris too. What's the furthest you've ever traveled? \\
\textcolor{blue}{Alice} & I went to Spain with my family last summer! \\
\textcolor{red}{Bob} & Wow, that's really cool! I've never been out of the country either. \\
\textcolor{orange}{Claire} & You should definitely go to Paris! What do you like to do for fun? \\
\textcolor{red}{Bob} & Well, I like to read sci-fi books, and I love to play board games. \\
\textcolor{blue}{Alice} & Do you have a favorite board game? \\
\textcolor{red}{Bob} & My favorite game is Settlers of Catan. My friends and I play it all the time. \\
\textcolor{orange}{Claire} & Haha, I love Settlers of Catan! My family and I play it all the time too. \\
\textcolor{blue}{Alice} & That's funny, I love it too. I'm in a Facebook group for it. \\
\bottomrule
\end{tabular}
\caption{Synthetic triadic conversation generated by OPT 30B which was rated as completely incoherent. First, ``Alice'' mentions they went on a road trip around California, then they mention that they haven't been out of the country. This implies that ``Alice'' is located in the United States. But then Alice mentions having traveled to Spain. The conversation also randomly shifts from discussing travel plans to discussing board games and other hobbies with an awkwardly placed transition on ``Claire's'' third turn.}
\label{ErrorAnalysis9}
\end{table*}

\begin{table*}[!htbp]
\small
\begin{tabular}{p{0.1\linewidth}p{0.8\linewidth}}
Party & Utterance \\ \hline
\textcolor{blue}{Alice} & Hello! How's your garden doing? \\
\textcolor{orange}{Claire} & It's doing great! I have a young garden, so I'm still waiting for it to develop. \\
\textcolor{blue}{Alice} & I can't wait to get home and check on mine! What are you growing? \\
\textcolor{orange}{Claire} & I'm growing tomatoes, strawberries, watermelon, and sunflowers! \\
\textcolor{blue}{Alice} & That sounds really nice! Do you have a garden somewhere else too? \\
\textcolor{orange}{Claire} & No, this is my first garden! \\
\textcolor{blue}{Alice} & Oh, I'm jealous! I would love to have my own garden someday. \\
\textcolor{red}{Bob} & I bet you would! I bet you would have a green thumb too. \\
\textcolor{blue}{Alice} & Maybe! Maybe I will try starting a garden next year! \\
\bottomrule
\end{tabular}
\caption{Synthetic triadic conversation generated by OPT 30B which was rated as completely unnatural. The conversation appears to be coherent, but the speakers appear to be very enthusiastic. It could be argued that the speakers are overly unenthusiastic to an unnatural degree.}
\label{ErrorAnalysis10}
\end{table*}

\begin{table*}[!htbp]
\small
\scalebox{0.9}{
\begin{tabular}{p{0.1\linewidth}p{0.8\linewidth}}
Party & Utterance \\ \hline
\textcolor{blue}{Alice} & You know, I think the EU is just a big conspiracy. \\
\textcolor{red}{Bob} & Conspiracy? \\
\textcolor{blue}{Alice} & Yes, the European Union wants to take over all of the land from its original owners. \\
\textcolor{red}{Bob} & What do you mean? \\
\textcolor{blue}{Alice} & I'm sure that's what they would tell you. But the truth is they're enslaving the people to gain control over resources. \\
\textcolor{orange}{Claire} & It sounds a bit extreme. How do you feel about EU leaders? \\
\textcolor{red}{Bob} & I don't like them. \\
\textcolor{blue}{Alice} & They're not very nice, I agree. \\
\textcolor{red}{Bob} & I'm kind of scared to go to EU countries and I feel like I don't even have any power. \\
\textcolor{blue}{Alice} & Oh, but we do! They won't take away the land of the original owners, because that's just not going to happen! \\
\textcolor{red}{Bob} & I'm still afraid. \\
\textcolor{orange}{Claire} & Hm. That sounds like a pretty horrible situation you're in. \\
\textcolor{blue}{Alice} & It is! And I don't know what to do. \\
\textcolor{red}{Bob} & Well, it sounds like we all have a lot to learn. I don't know what we should do, but I think there are lots of options out there. \\
\textcolor{blue}{Alice} & It's nice that you have an open mind about this! I really do think people should learn more about the EU and what's going on. \\
\bottomrule
\end{tabular}
}
\caption{Synthetic triadic conversation generated by OPT 30B which was rated as generic and dull. ``Alice'' begins a long debate on whether the EU is a ``conspiracy'' without making a lot of conversational progress.}
\label{ErrorAnalysis12}
\end{table*}

\begin{table*}[!htbp]
\scalebox{0.9}{
\begin{tabular}{p{0.05\linewidth}p{0.95\linewidth}}
Party & Utterance \\ \hline
john & sure \\
john & i think so, meg \\
mara & how did i know that was coming \\
meg & its not just the public eye \\
john & haha  mara \\
mara & hushh.. *** \\
nick & There are already other countries who are investigating the Bush administration for war crimes - Spain \\
meg & with the breton woods \\
george & they need to be prosecuted...that's in obama's hands \\
\midrule
nick & wow, george, right win propaganda... huh \\
meg & look at how well Iraq is doing \\
mara & goodness \\
meg & there's a point at which interrogation becomes torture and is just inhumane \\
john & agree to george \\
mara & ? \\
mara & im in albany btw \\
meg & Which we signed! \\
\midrule
amy & well it is the way the world is going-- email, chat,, etc \\
john & yes \\
jordan & And this is one of the tricky things in this virtual world. You know nothing about the people u r talking to!!!! \\
amy & u r right   you just used online language   haha \\
mara & hes not much fun either haha, what do you think? \\
amy & hi john-  can you see my message here? \\ 
jordan & Hi, amy \\
mara &  i dont know what is better really!!! \\
john & haha \\
\bottomrule
\end{tabular}
}
\caption{Three excerpts of the same conversation from the MPC corpus~\cite{shaikh2010mpc}. The conversation spans topics ranging from the Bush administration to meta-discussion about the collection task.}
\label{MPCExample1}
\end{table*}

\begin{table*}[!htbp]
\begin{tabular}{p{0.15\linewidth}p{0.85\linewidth}}
Party & Utterance \\ \hline
Phoebe & Then I’m gonna have to ask you to keep it down. \\
Mr. Heckles & Who are you? \\
Eric & Hi, I’m Eric, I’m gonna be Chandler’s new roommate. \\
Mr. Heckles & I’m Chandler’s new roommate. \\
Eric & I-I-I don’t think so. \\
Mr. Heckles & I could be Chandler’s new roommate. \\
Eric & But, he told me over the phone. \\
Mr. Heckles & He told me in person. \\
Eric & That’s weird. \\
Mr. Heckles & Well, I’m going to go into my new apartment now.  Ehh! \\
\bottomrule
\end{tabular}
\caption{Conversation from the MELD corpus~\cite{poria2019meld}. Three speakers are involved, discussing a living situation regarding a fourth character who does not appear in this scene.}
\label{MELDExample}
\end{table*}

\begin{table*}[!htbp]
\small
\begin{tabular}{p{0.2\linewidth}p{0.75\linewidth}}
Subtopic & Background Information \\ \hline
Pacific Theater & Alice is interested in Pacific theater. \\
Growing residential grass & Alice is interested in growing residential grass. \\
Breakfast food & Alice likes to try different breakfast foods. Bob loves waffles. \\
music & Alice likes music. Bob plays the viola. \\
skincare & Alice is interested in skincare. Bob has a great skincare routine. \\
Planting flowers & Alice is interested in planting flowers. Bob has a nice garden. \\
Southern Ice Tea & Alice is interested in Southern Ice Tea. Bob has a great recipe. \\
herb garden & Alice is interested in planting an herb garden. \\
Hiking & Alice is going hiking tomorrow. \\
Plant a garden & Alice wants to plant a garden. \\
Italian food & Alice likes Italian food. \\
book recommendations & Alice is interested in book recommendations. \\
anniversaries & Alice keeps track of all of her anniversaries. \\
Existential Psychology & Alice is interested in Existential Psychology. \\
The Outlander Series & Alice is interested in The Outlander Series. \\
camping gear & Alice is looking for advice on camping gear. Bob works at REI. \\
Movie & Alice is interested in movie recommendations. Bob is a film buff. \\
Ford Vehicles & Alice is interested in Ford vehicles. Bob prefers Japanese cars. \\
Beauty & Alice is interested in beauty. Bob works at Sephora. \\
Syrian War & Alice is interested in the Syrian War. Bob is a political scientist. \\
Elon Musk & Alice and Bob are talking about Elon Musk. \\
Healthy foods & Alice and Bob are discussing healthy foods. Alice is on a paleo diet. \\
Soren Kierkegaard & Alice is a fan of Soren Kierkegaard. \\
investing money & Alice is interested in investing money. Bob is an investment banker. \\
Post-structuralism & Alice is interested in post-structuralism. \\
baking & Alice is interested in baking. Bob has baked cakes and brownies before. \\
Nuts & Alice likes to eat nuts. \\
braids & Alice braids her hair. Bob is interested in learning how. \\
Growing vegetables & Alice is interested in growing vegetables. \\
Martin Luther & Alice is learning about Martin Luther. \\
paint brushes & Alice is interested in paint brushes. \\
Stock Trading & Alice is interested in stock trading. \\
Install TV applications & Alice wants to install TV applications. Bob is helping her. \\
History & Alice is interested in history. History was Bob's favorite school subject. \\
Feminism & Alice is interested in feminism. Bob majored in gender studies. \\
Tell a joke & Alice wants to hear Bob tell a joke. \\
artists & Alice is interested in learning about modern artists. \\
Turtles & Alice likes turtles. Bob has been scuba diving. \\
Anthony Trollope & Alice likes the work of Anthony Trollope. Bob prefers modern literature. \\
Paris & Alice wants to go to Paris. \\
Bread & Alice likes bread. Bob's favorite bread is a baguette. \\
movie cast members & Alice and Bob are talking about movie cast members. \\
Gay Marriage & Alice is a proponent of gay marriage. Bob is interested in learning more. \\
U.S. Senate & Alice and Bob are discussing the U.S. Senate. \\
growing tomatoes & Alice is interested in growing tomatoes. \\
family issues & Alice is interested in family issues. \\
Automotive parts & Alice is interested in automative parts. \\
Bee life & Alice is interested in bee life. \\
Taylor Swift & Alice's favorite musician is Taylor Swift. Bob likes Ariana Grande. \\
biking & Alice's favorite hobby is biking. Bob prefers rock climbing. \\
Juicers & Alice wants to get a juicer. \\
islands & Alice likes visiting islands. Bob prefers hiking. \\
Planets & Alice is learning about the planets in school. \\
Pokemon & Alice likes to play Pokemon. Bob also likes Pokemon. \\
\bottomrule
\end{tabular}
\caption{Corresponding background information written for each of the subtopics found in the FITS dataset. There is a mixture of prompts which only mention one speaker and prompts which mention two speakers. Every synthetic conversation involves both speakers.}
\label{DyadicPrompts}
\end{table*}

\begin{table*}[!htbp]
\small
\begin{tabular}{p{0.2\linewidth}p{0.75\linewidth}}
Topic & Conversation Recipe \\ \hline
Growing residential grass & Alice is interested in growing residential grass. Claire has a really neat yard. \\
Breakfast food & Alice likes to try different breakfast foods. Bob loves waffles. Claire prefers pancakes. \\
music & Alice likes music. Bob plays the viola. Claire played the violin in high school. \\
skincare & Alice is interested in skincare. Bob has a great skincare routine. Claire wants to hear Bob's routine. \\
Planting flowers & Alice is interested in planting flowers. Bob has a nice garden. Claire has a vegetable garden. \\
Southern Ice Tea & Alice is interested in Southern Ice Tea. Bob has a great recipe. Claire loved trying Bob's Southern Ice Tea. \\
herb garden & Alice is interested in planting an herb garden. Claire has some gardening tips. \\
Hiking & Alice is going hiking tomorrow. Claire hates hiking. \\
Plant a garden & Alice wants to plant a garden. Claire has a greenroom. \\
Italian food & Alice likes Italian food. Claire prefers Asian food. \\
book recommendations & Alice is interested in book recommendations. Claire is a part of a book club. \\
anniversaries & Alice keeps track of all of her anniversaries. Claire is not well-organized. \\
Existential Psychology & Alice is interested in Existential Psychology. Claire is a psychologist by training. \\
The Outlander Series & Alice is interested in The Outlander Series. Claire has never seen the series. \\
camping gear & Alice is looking for advice on camping gear. Bob works at REI. Claire loves the outdoors. \\
Movie & Alice is interested in movie recommendations. Bob is a film buff. Claire is also a film buff. \\
Ford Vehicles & Alice is interested in Ford vehicles. Bob prefers Japanese cars. Claire prefers to drive a BMW. \\
Beauty & Alice is interested in beauty. Bob works at Sephora. Claire is shopping with Alice. \\
Syrian War & Alice is interested in the Syrian War. Bob is a political scientist. Claire is studying modern political theory. \\
Elon Musk & Alice and Bob are talking about Elon Musk. Claire is a Tesla owner. \\
Healthy foods & Alice and Bob are discussing healthy foods. Alice is on a paleo diet. Claire is a nutritionist. \\
Soren Kierkegaard & Alice is a fan of Soren Kierkegaard. Claire is not familiar with Soren Kierkegaard. \\
investing money & Alice is interested in investing money. Bob is an investment banker. CLaire is an expert in personal finance. \\
Post-structuralism & Alice is interested in post-structuralism. Claire is an expert on the subject. \\
baking & Alice is interested in baking. Bob has baked cakes and brownies before. Claire wants to learn how to bake. \\
Nuts & Alice likes to eat nuts. Claire is allergic to peanuts. \\
braids & Alice braids her hair. Bob is interested in learning how. Claire braids her hair every day. \\
Growing vegetables & Alice is interested in growing vegetables. Claire has a vegetable garden. Bob grows flowers. \\
Martin Luther & Alice is learning about Martin Luther. Claire is a historian. \\
paint brushes & Alice is interested in paint brushes. Claire is a painter and has several suggestions. \\
Stock Trading & Alice is interested in stock trading. Claire is a stock broker. \\
Install TV applications & Alice wants to install TV applications. Bob is helping her. Claire is also good with technology. \\
History & Alice is interested in history. History was Bob's favorite school subject. Claire is a historian. \\
Feminism & Alice is interested in feminism. Bob majored in gender studies. Claire does not know much about feminism. \\
Tell a joke & Alice wants to hear Bob tell a joke. Claire is a stand-up comedian. \\
artists & Alice is interested in learning about modern artists. Claire is a photographer. \\
Turtles & Alice likes turtles. Bob has been scuba diving. Claire wants to try scuba diving. \\
Anthony Trollope & Alice likes the work of Anthony Trollope. Bob prefers modern literature. Claire is not familiar with much literature. \\
Paris & Alice wants to go to Paris. Claire has never been to Europe. \\
Bread & Alice likes bread. Bob's favorite bread is a baguette. Claire loves to bake bread. \\
movie cast members & Alice and Bob are talking about movie cast members. Claire has seen a lot of movies recently. \\
Gay Marriage & Alice is a proponent of gay marriage. Bob is interested in learning more. Claire is an activist. \\
U.S. Senate & Alice and Bob are discussing the U.S. Senate. Claire is a politician. \\
growing tomatoes & Alice is interested in growing tomatoes. Claire has a large garden with many tomatoes. \\
family issues & Alice is interested in family issues. Claire is a therapist. \\
Automotive parts & Alice is interested in automative parts. Claire is a mechanic. \\
Bee life & Alice is interested in bee life. Claire is a beekeeper. \\
Taylor Swift & Alice's favorite musician is Taylor Swift. Bob likes Ariana Grande. Claire does not like pop music. \\
biking & Alice's favorite hobby is biking. Bob prefers rock climbing. Claire prefers archery. \\
Juicers & Alice wants to get a juicer. Claire has a suggestion for a great juicer. \\
islands & Alice likes visiting islands. Bob prefers hiking. Claire likes the beach. \\
Planets & Alice is learning about the planets in school. Claire is an astronomer. \\
Pokemon & Alice likes to play Pokemon. Bob also likes Pokemon. Claire prefers to play Stardew Valley. \\
\bottomrule
\end{tabular}
\caption{Triadic background information written for each of the subtopics given in the FITS dataset. Unlike Table~\ref{DyadicPrompts}, each of these may include background information for up to three people.}
\label{TriadicPrompts}
\end{table*}

\begin{table*}[!htbp]
\tiny
\begin{tabular}{p{0.9\linewidth}} \toprule
The following is a conversation between Alice and Bob about past travel experiences. Alice has been to Japan and Bob is considering flying there.\\
Alice: Hi!\\
Bob: Hey, how are you doing?\\
Alice: I'm doing well! I just got back from my vacation in Japan.\\
Bob: Wow that's awesome! What did you think of it?\\
Alice: Japan was such an amazing place to visit!\\
Bob: Wow! What was your favorite part?\\
Alice: I really enjoyed the food in Tokyo.\\
Bob: Which airline did you take?\\
Alice: I flew using Japan Airlines.\\
\midrule
The following is a conversation between Alice and Bob about their hobbies. Alice enjoys tennis and Bob likes playing soccer.\\
Alice: What do you like to do for fun?\\
Bob: I used to play soccer in college, so I still like to play for fun on the weekends!\\
Alice: That's great. Soccer is a great way to stay in good shape.\\
Bob: I agree - it's really good cardio. What about you?\\
Alice: I love to play tennis. I've been taking lessons for a few months now!\\
Bob: Tennis is fun too! \\
\midrule
The following is a conversation between Alice and Bob about their favorite movies. Bob loved the new Batman movie. Alice really liked watching Pride and Prejudice.\\
Alice: I just saw Pride and Prejudice for the fifth time!\\
Bob: That's a lot of times! What do you like so much about that movie?\\
Alice: Well, as a teenager I really liked the book. But I just really loved Keira Knightley's portrayal of Elizabeth.\\
Bob: I see. I haven't seen the movie myself. I prefer action films.\\
Alice: What's your favorite action movie?\\
Bob: Hm, I really liked the Batman movie that just came out.\\
Alice: I haven't seen it yet. I heard it got pretty good reviews. \\
\midrule
The following is a conversation between Alice and Bob about their hometowns. Alice is from New York City. Bob grew up in Seattle.\\
Alice: Hello! How are you doing?\\
Bob: Hi, I'm doing great! What about yourself?\\
Alice: I'm doing well! Where are you from?\\
Bob: I'm originally from Seattle, but now I live in Palo Alto.\\
Alice: Oh cool! I live in Palo Alto too. Do you like Seattle or California more?\\
Bob: Well, Seattle is always going to be home for me. Even if the weather in California is nicer.\\
Alice: Haha, I get that! I miss New York City - there's no place like home.\\
Bob: What is your favorite neighborhood of New York City?\\
Alice: I love going to Chelsea. The Highline has a great view, and Little Island is close by too! Have you ever been?\\
Bob: Unfortunately I have not. I have never been to the East Coast! \\
\midrule
The following is a conversation between Alice and Bob about art. Alice's favorite artist is Michelangelo. Bob does not know much about art.\\
Alice: Hi, how's it going?\\
Bob: It's going well, what about you?\\
Alice: I'm doing great! I've been really interested in art recently.\\
Bob: What got you interested in art?\\
Alice: Art can be so breathtaking!\\
Bob: I feel like I don't know how to properly appreciate art, but certain pieces of artwork certainly look very complex.\\
Alice: Have you ever heard of Michelangelo?\\
Bob: I have heard of him, but I don't know anything that he has created.\\
Alice: Michelangelo is really famous for his statue of David.\\
Bob: Huh? Who is David?\\
Alice: David is a Biblical figure who was a king of Israel. Michelangelo built a really magnificent statue of him in Florence. \\
\midrule
The following is a conversation between Alice and Bob about drinks. Alice is a wine expert, whereas Bob prefers cocktails.\\
Alice: How are you doing?\\
Bob: Pretty great! I'm planning to go to a brewery this weekend.\\
Alice: Do you know much about alcohol?\\
Bob: Yeah, I really like beer! I drink a lot of IPAs.\\
Alice: Oh - what do you like about IPAs? I can't get over the bitter taste.\\
Bob: Well, I don't think it's just bitter. Sometimes there are really interesting citrusy or herbal flavor notes.\\
Alice: I see. That kind of reminds me of wine tasting.\\
Bob: There's definitely a lot of depth to it like there is with wine. Do you know much about wine?\\
Alice: Yeah, I took several classes on wine tasting back in the day. I really love Pinot Noir.\\
Bob: Oh I love red wines too.\\
Alice: Right? I love the dryness and fruity notes of Pinot Noir. \\
\midrule
The following is a conversation between Alice and Bob about relationships. Bob recently got engaged.\\
Alice: Congrats on your engagement! When do you think you will have your wedding?\\
Bob: Thank you!! We're thinking of having it in November.\\
Alice: That's amazing! Will you pick a fancy destination?\\
Bob: I wanted to! I was thinking of having it somewhere in Europe, but my partner and I ultimately decided we wanted to have it close to home so our friends could all make it.\\
Alice: That's a good point. My husband and I had similar thoughts when we were planning our wedding.\\
Bob: What did you plan in the end?\\
Alice: We had a small ceremony in my hometown! \\
\midrule
The following is a conversation between Alice and Bob about their jobs. Alice works in the financial industry and Bob is a musician.\\
Alice: I'm so burnt out from my work! I just want to quit already!\\
Bob: Whoa - what do you do for work? \\
Alice: I'm an investment banker. It's been four years at this company and I'm absolutely exhausted.\\
Bob: That sounds intense. Is there anything you actually like about the job?\\
Alice: Well, the money is good.\\
Bob: It sounds like you could use a break. Maybe you could use some of that money to go travel.\\
Alice: I really want to go to South America, but I don't have a lot of time. \\
\midrule
The following is a conversation between Alice and Bob about their pets. Alice has a dog and Bob prefers cats.\\
Alice: Do you have any pets?\\
Bob: No, but I really want to get a cat.\\
Alice: What, why a cat? Cats seem so boring. They never want to play.\\
Bob: Yeah, but cats are so cute! They also are a lot easier to take care of. They can clean themselves. What do you prefer?\\
Alice: Well, I have a dog. He is a corgi and his name is Bo.\\
Bob: Aww that's cute! I'm not usually a dog person, but corgis are adorable.\\
Alice: Haha, thank you! Bo is a really friendly dog.\\
Bob: How old is he?\\
Alice: Bo is one year old now. \\
\midrule
The following is a conversation between Alice and Bob about grocery shopping. Alice has a shopping list for Bob.\\
Alice: Could you run to the grocery store and pick up some bananas for me?\\
Bob: Will do - how many do you need?\\
Alice: Oh, I don't know, maybe ten bananas. I'm planning to make banana bread, but I also want to save some for us to eat at home.\\
Bob: That sounds delicious! I'll head out in a second. Is there anything else you need? \\
\bottomrule
\end{tabular}
\caption{Handwritten conversation examples of varying length. In-context examples are randomly sampled from this pool and used as part of a prompt for dyadic conversation generation.}
\label{HWExamples}
\end{table*}

\begin{table*}[!htbp]
\tiny
\begin{tabular}{p{0.99\linewidth}}
\toprule
The following is a conversation between Alice and Bob and Claire about past travel experiences. Alice has been to Japan and Bob is considering flying there. Claire has been to Taiwan and Korea, but not Japan.\\
Alice: Hi!\\
Bob: Hey, how are you doing?\\
Alice: I'm doing well! I just got back from my vacation in Japan.\\
Bob: Wow that's awesome! What did you think of it?\\
Alice: Japan was such an amazing place to visit!\\
Claire: Wow, I've always wanted to visit Japan!\\
Bob: What was your favorite part?\\
Alice: I really enjoyed the food in Tokyo. I had the best sushi of my life!\\
Bob: Which airline did you take?\\
Alice: I flew using Japan Airlines.\\
Claire: How expensive are tickets these days?\\

\midrule
The following is a conversation between Alice and Bob about their hobbies. Alice enjoys tennis and Bob likes playing soccer. Claire plays football.\\
Alice: What do you like to do for fun?\\
Bob: I used to play soccer in college, so I still like to play for fun on the weekends!\\
Claire: Oh wow! Did you play varsity soccer?\\
Bob: Yeah, I was a four-year starter!\\
Alice: That's great. Soccer is a great way to stay in good shape.\\
Bob: I agree - it's really good cardio. What about you all?\\
Claire: I'm in a flag football league! We play every Saturday afternoon.\\
Alice: I love to play tennis. I've been taking lessons for a few months now!\\
Bob: Cool, football and tennis are fun too! \\
\midrule
The following is a conversation between Alice and Bob and Claire about their favorite movies. Claire is looking for movie recommendations. Bob loved the new Batman movie. Alice really liked watching Pride and Prejudice.\\
Alice: I just saw Pride and Prejudice for the fifth time!\\
Claire: Would you recommend watching it? I've never seen it!\\
Bob: Yeah, five times is a lot of times! What do you like so much about that movie?\\
Alice: Well, as a teenager I really liked the book. But I just really loved Keira Knightley's portrayal of Elizabeth.\\
Bob: I see. I haven't seen the movie myself. I prefer action films.\\
Alice: What's your favorite action movie?\\
Bob: Hm, I really liked the Batman movie that just came out.\\
Alice: I haven't seen it yet. I heard it got pretty good reviews.\\
\midrule
The following is a conversation between Alice and Bob and Claire about their hometowns. Alice is from New York City. Bob grew up in Seattle. Claire is from Boston and would like to visit New York City.\\
Alice: Hello! How are you doing?\\
Claire: I'm doing good!\\
Bob: Hi, I'm doing great! What about yourself?\\
Alice: I'm doing well! Where are you both from?\\
Claire: I'm from Boston! I'm just visiting the Bay Area.\\
Bob: I'm originally from Seattle, but now I live in Palo Alto.\\
Alice: Oh cool! I live here in Palo Alto. Do you like Seattle or California more?\\
Bob: Well, Seattle is always going to be home for me. Even if the weather in California is nicer.\\
Alice: Haha, I get that! I miss New York City - there's no place like home.\\
Claire: Oh you're from New York? I've always wanted to visit!\\
Bob: Me too! What is your favorite neighborhood of New York City?\\
Alice: I love going to Chelsea. The Highline has a great view, and Little Island is close by too! Have you ever been?\\
Bob: Unfortunately I have not. I have never been to the East Coast!\\
\midrule
The following is a conversation between Alice and Bob and Claire about art. Alice's favorite artist is Michelangelo. Bob does not know much about art. Claire is a painter.\\
Alice: Hi, how's it going?\\
Bob: It's going well, what about you?\\
Alice: I'm doing great! I've been really interested in art recently.\\
Claire: Oh that's great to hear! I love art as well.\\
Bob: What got you interested in art?\\
Alice: Art can just be so breathtaking!\\
Bob: I feel like I don't know how to properly appreciate art, but certain pieces of artwork certainly look very complex.\\
Alice: Have you ever heard of Michelangelo?\\
Bob: I have heard of him, but I don't know anything that he has created.\\
Claire: Michelangelo has some truly magnificent paintings, such as The Creation of Adam.\\
Alice: Michelangelo is also really famous for his statue of David.\\
Bob: Huh? Who is David?\\
Alice: David is a Biblical figure who was a king of Israel. Michelangelo built a really magnificent statue of him in Florence.\\
\midrule
The following is a conversation between Alice and Bob and Claire about drinks. Alice is a wine expert, whereas Bob prefers cocktails. Claire likes to drink beer.\\
Alice: How are you doing?\\
Bob: Pretty great! I'm planning to go to a brewery this weekend.\\
Alice: Do you know much about alcohol?\\
Bob: Yeah, I really like beer! I drink a lot of IPAs.\\
Claire: Oh, beers are my favorite type of drink! I can really appreciate the taste of a good IPA.\\
Alice: Oh - what do you like about IPAs? I can't get over the bitter taste.\\
Bob: Well, I don't think it's just bitter. Sometimes there are really interesting citrusy or herbal flavor notes.\\
Claire: Yeah, there's a whole science to the hops used in making IPAs!\\
Alice: I see. That kind of reminds me of wine tasting.\\
Claire: The science behind tasting is similar for sure.\\
Bob: I agree, there's definitely a lot of depth to it like there is with wine. Do you know much about wine?\\
Alice: Yeah, I took several classes on wine tasting back in the day. I really love Pinot Noir.\\
Bob: Oh I love red wines too.\\
Alice: Right? I love the dryness and fruity notes of Pinot Noir.\\
\bottomrule
\end{tabular}
\caption{Triadic conversation recipes written for each of the ``generic topics'' given in the FITS dataset. These conversation recipes are included after the in-context examples when prompting PLMs to generate synthetic conversations. Unlike Table~\ref{DyadicPrompts}, each of these conversation recipes may include background for up to three people. Continued in Table~\ref{HWTriadicExamplesPt2}.}
\label{HWTriadicExamples}
\end{table*}

\begin{table*}[!htbp]
\small
\begin{tabular}{p{0.99\linewidth}}
\toprule
The following is a conversation between Alice and Bob and Claire about relationships. Bob recently got engaged.\\
Alice: Congrats on your engagement! \\
Claire: Yes, congrats! When do you think you will have your wedding?\\
Bob: Thank you! We're thinking of having it in November.\\
Alice: That's amazing! Will you pick a fancy destination?\\
Bob: I wanted to! I was thinking of having it somewhere in Europe, but my partner and I ultimately decided we wanted to have it close to home so our friends could all make it.\\
Claire: Oh wow, that is very considerate of you.\\
Alice: Yeah, that's a good point. My husband and I had similar thoughts when we were planning our wedding.\\
Bob: What did you plan in the end?\\
Alice: We had a small ceremony in my hometown!\\
Claire: It turned out nicely! It was such a beautiful ceremony. \\
\midrule
The following is a conversation between Alice and Bob and Claire about their jobs. Alice works in the financial industry and Bob is a musician. Claire is an architect.\\
Alice: I'm so burnt out from my work! I just want to quit already!\\
Bob: Whoa - what do you do for work? \\
Alice: I'm an investment banker. It's been four years at this company and I'm absolutely exhausted.\\
Bob: That sounds intense. Is there anything you actually like about the job?\\
Alice: Well, the money is good.\\
Claire: That doesn't sound like a healthy relationship with your job!\\
Bob: It sounds like you could use a break. Maybe you could use some of that money to go travel.\\
Alice: I really want to go to South America, but I don't have a lot of time.\\
Claire: Don't you have vacation days? I think breaks are important.\\
Alice: Yes, but I really want to get promoted this year. \\
\midrule
The following is a conversation between Alice and Bob and Claire about their pets. Alice has a dog and Bob prefers cats. Claire has a pet hamster.\\
Alice: Do you have any pets?\\
Claire: I have a pet hamster! He is so adorable. What about you two?\\
Bob: I don't, but I really want to get a cat.\\
Alice: What, why a cat? Cats seem so boring. They never want to play.\\
Bob: Yeah, but cats are so cute! They also are a lot easier to take care of. They can clean themselves. What do you prefer?\\
Alice: Well, I have a dog. He is a corgi and his name is Bo.\\
Claire: That's so adorable! How old is he?\\
Alice: He just turned one!\\
Bob: Aww that's cute! I'm not usually a dog person, but corgis are adorable.\\
Alice: Haha, thank you! Bo is a really friendly dog. \\
\midrule
The following is a conversation between Alice and Bob and Claire about grocery shopping. Alice has a shopping list for Bob. Claire is helping Alice cook at home.\\
Alice: Could you run to the grocery store and pick up some bananas for me?\\
Bob: Will do - how many do you need?\\
Alice: Oh, I don't know, maybe ten bananas. We are planning to make banana bread, but I also want to save some for us to eat at home.\\
Bob: That sounds delicious! I'll head out in a second. Is there anything else you need?\\
Claire: Oh, could you also pick up some more eggs? I think we're running low here. \\
\bottomrule
\end{tabular}
\caption{Triadic conversation recipes written for each of the ``generic topics'' given in the FITS dataset continued from Table~\ref{HWTriadicExamples}.}
\label{HWTriadicExamplesPt2}
\end{table*}
\end{document}